\documentclass[runningheads]{llncs}

\usepackage{eccv}

\usepackage{color}
\usepackage{xcolor,colortbl}

\usepackage{eccvabbrv}

\usepackage{graphicx}
\usepackage{booktabs}
\usepackage{multirow}

\usepackage[normalem]{ulem}
\useunder{\uline}{\ul}{}

\newcommand*\samethanks[1][\value{footnote}]{\footnotemark[#1]}

\usepackage[accsupp]{axessibility}  

\usepackage[pagebackref,breaklinks,colorlinks,citecolor=eccvblue]{hyperref}

\usepackage{orcidlink}

\begin{document}

\title{CoR-GS: Sparse-View 3D Gaussian Splatting via Co-Regularization} 

\titlerunning{CoR-GS}


\author{Jiawei Zhang\inst{1} \and
Jiahe Li \inst{1} \and
Xiaohan Yu \inst{3} \and
Lei Huang \inst{2} \and
Lin Gu\inst{4,5} \and
Jin Zheng\inst{1}\thanks{Corresponding author: Xiao Bai and Jin Zheng} \and
Xiao Bai\inst{1}\samethanks \\
}

\authorrunning{J. Zhang et al.}


\institute{School of Computer Science and Engineering, State Key Laboratory of \\ Complex \& Critical Software Environment, Jiangxi Research Institute, \\ Beihang University \and SKLCCSE, Institute of Artificial Intelligence, Beihang University \and School of Computing, Macquarie University \and RIKEN AIP \and 
The University of Tokyo}

\maketitle

\begin{abstract}
  3D Gaussian Splatting (3DGS) creates a radiance field consisting of 3D Gaussians to represent a scene. With sparse training views, 3DGS easily suffers from overfitting, negatively impacting rendering. This paper introduces a new co-regularization perspective for improving sparse-view 3DGS. When training two 3D Gaussian radiance fields, we observe that the two radiance fields exhibit point disagreement and rendering disagreement that can unsupervisedly predict reconstruction quality, stemming from the randomness of densification implementation. We further quantify the two disagreements and demonstrate the negative correlation between them and accurate reconstruction, which allows us to identify inaccurate reconstruction without accessing ground-truth information. Based on the study, we propose CoR-GS, which identifies and suppresses inaccurate reconstruction based on the two disagreements: (\romannumeral1) Co-pruning considers Gaussians that exhibit high point disagreement in inaccurate positions and prunes them. (\romannumeral2) Pseudo-view co-regularization considers pixels that exhibit high rendering disagreement are inaccurate and suppress the disagreement. Results on LLFF, Mip-NeRF360, DTU, and Blender demonstrate that CoR-GS effectively regularizes the scene geometry, reconstructs the compact representations, and achieves state-of-the-art novel view synthesis quality under sparse training views. Project page: \url{https://jiaw-z.github.io/CoR-GS}
\end{abstract}

\section{Introduction}
\label{sec:intro}

Obtaining 3D representations from 2D images has long been a topic of interest. 3D Gaussian Splatting (3DGS) \cite{kerbl20233d} creates unstructured radiance fields consisting of a set of 3D Gaussians to represent the scene and has achieved high-quality novel view synthesis in real time. However, when only sparse views are available, the decrease in training constraints makes 3DGS prone to overfit training views \cite{zhu2023FSGS,Xiong2023sparsegs}, resulting in unrealistic novel view synthesis.

3DGS initializes 3D Gaussians with a sparse point cloud. Then an interleaved optimization/density control of 3D Gaussians is performed to achieve an accurate radiance field representation of the scene. Under sparse training views, we observe that although two 3D Gaussian radiance fields are trained to represent the same scene, they exhibit differences in both Gaussian positions and rendered pixels. Their differences significantly increase during density control, which involves creating new Gaussians and initializing their positions by sampling from a normal distribution. With sparse training views, the optimization can struggle to correct Gaussians to accurately represent the scene due to the ambiguities of 3D to 2D projection, resulting in an accumulation of differences.

\begin{figure}[t!]
    \centering
    \includegraphics[width=1.0\linewidth]{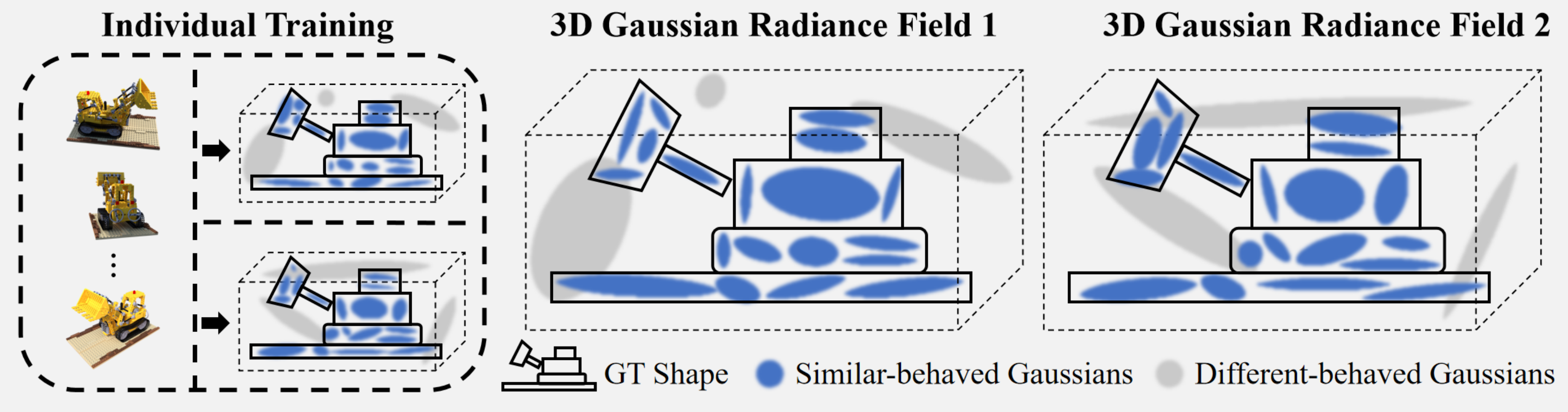}
    \caption{Illustration of how the different behaviors between two 3D Gaussian radiance fields correlated to construction quality. Gaussians with different behaviors tend to not fit the ground-truth shape well. Therefore, inaccurate reconstructions can be identified by measuring the differences without accessing ground-truth information.}
    \label{fig:intro_overview}
\end{figure}

In this paper, we reveal the link between the different behaviors of two 3D Gaussian radiance fields and their reconstruction quality. Specifically, we propose point disagreement and rendering disagreement to indicate the differences and measure them quantitatively. The point disagreement indicates the differences in the Gaussian position, which is evaluated on the registration between Gaussians' point cloud representations. The rendering disagreement indicates the differences in their rendered pixels. We compare the rendered images to ground-truth test views to measure the reconstruction quality of 3D Gaussian radiance fields. We also utilize a 3D Gaussian field trained with dense views as ground truth to evaluate Gaussian positions and rendered depth maps to provide a more comprehensive assessment of reconstruction quality. More details are provided in \cref{sec:empirical study}. The experimental results demonstrate a negative correlation between the two disagreements and accurate reconstruction. This allows us to unsupervisedly identify inaccurate reconstruction by comparing two 3D Gaussian radiance fields, as illustrated in \cref{fig:intro_overview}.

Based on our study, we propose CoR-GS, which trains two 3D Gaussian radiance fields with the same views and conducts co-regularization during training. It improves sparse-view 3DGS by identifying and suppressing inaccurate reconstruction based on the point disagreement and rendering disagreement. CoR-GS implies co-pruning to suppress point disagreement. Co-pruning treats two 3D Gaussian radiance fields as two point clouds, performing point-wise matching between them. It considers Gaussians that do not have nearby matching points in the opposite point cloud as outliers and prunes them. To suppress rendering disagreement, CoR-GS implies pseudo-view co-regularization. It samples online pseudo views by interpolating training views and considers pixels that exhibit high rendering disagreement are inaccurately rendered. To suppress the inaccurate rendered results, it computes the differences of rendered pixels as a regularization term combined with the training-view loss. 

Integrating co-pruning and pseudo-view co-regularization, CoR-GS reconstructs coherent and compact geometry and achieves state-of-the-art sparse-view rendering performance on LLFF, Mip-NeRF360, DTU, and Blender datasets. The experiments demonstrate our method's universal ability to regularize sparse-view 3DGS in various scene situations.

Our main contributions are the following:
\begin{itemize}
    \item We propose point disagreement and rendering disagreement to measure the differences between two 3D Gaussian radiance fields for the same scene and demonstrate the negative correlation between the two kinds of disagreement and accurate reconstruction. The two agreements can be used to access the reconstruction quality without ground-truth information.
    \item We propose co-pruning and pseudo-view co-regularization for suppressing the point disagreement and rendering disagreement, respectively. We demonstrate that suppressing the two disagreements leads to more accurate 3D Gaussian radiance fields for representing the scene.
    \item Equipped with co-pruning and pseudo-view co-regularization, CoR-GS reconstructs coherent and compact geometry and achieves competitive quality across multiple benchmarks compared to the state-of-the-art methods.
\end{itemize}

\noindent We hope the observations and discussions in this paper can stimulate further thinking on the randomness of 3D Gaussian radiance fields.

\section{Related Work}
\label{sec:related}

\subsection{Radiance Fields} Radiance Fields are employed for reconstructing 3D scenes and synthesizing novel views. Neural Radiance Fields (NeRFs) \cite{mildenhall2021nerf} have experienced significant advancements, which learn the neural volumetric representations to represent 3D scenes and render images through volume rendering. Since then, a great number of studies aim to improve the rendering quality \cite{barron2021mip, barron2022mip360} and efficiency \cite{chen2022tensorf, muller2022instant, sun2022direct, yu2021plenoxels, yu2021plenoctrees, liu2020nsvf, hu2023tri} for NeRFs. Recent advancements in unstructured radiance fields, as demonstrated by studies like \cite{chen2023neurbf, xu2022point, kerbl20233d}, employ a collection of Gaussians to represent 3D scenes. Notably, 3D Gaussian Splatting (3DGS) \cite{kerbl20233d} represents scenes using a set of anisotropic 3D Gaussians, and renders images via differentiable splitting. 3DGS has demonstrated remarkable success in reconstructing high-quality complex scenes in real time. Despite the tremendous success of 3DGS in many 3D tasks \cite{luiten2023dynamic, wu20234d, tang2023dreamgaussian}, the behavior of 3DGS with sparse views is less investigated, posing an open problem in the field.

\subsection{Novel View Synthesis with Sparse Views} Novel view synthesis aims to generate unseen views of objects or scenes from a set of given images \cite{zhou2016view, avidan1997novel}. While given sparse views, many methods tend to reconstruct degraded scenes and render unrealistic novel views. Many studies have explored regularizing NeRFs with sparse views \cite{yang2023freenerf, niemeyer2022regnerf, kim2022infonerf, deng2022dsnerf}. Some methods focus on designing generative models and pre-train them on extensive datasets \cite{chen2021mvsnerf,yu2021pixelnerf,cong2023enhancing,zhou2023sparsefusion, kulhanek2022viewformer}. Conversely, others \cite{wynn2023diffusionerf, jain2021putting,deng2022dsnerf,roessle2022dense,song2023darf,wang2023sparsenerf} utilize the external knowledge from pre-trained models to regularize the training process.

With the advancements of 3DGS, recent studies \cite{zhu2023FSGS,Xiong2023sparsegs,li2024dngaussian} have observed the issue of 3DGS to synthesize unrealistic novel views in sparse-view scenarios. These methods rely on predictions of pre-trained depth estimators as regularization to correct the reconstructed geometry. However, external supervision can introduce additional noise, negatively impacting the reconstruction. Our paper introduces a new co-regularization perspective for sparse-view 3DGS by suppressing the disagreement between two 3D Gaussian radiance fields.

\subsection{Prediction Agreement} The agreement of predictions of two nerual networks has been explored in many tasks \cite{sindhwani2005co,blum1998combining,li2020dividemix,han2018co,wei2020combating,yu2019does,zhang2024robust}, especially for semi-supervised learning \cite{sindhwani2005co} and learning with noisy labels \cite{han2018co}. The typical pipeline involves simultaneously training two neural networks. They leverage the agreement of different networks' predictions to pseudo-label unlabeled data \cite{li2020dividemix} or clean noise from labeled annotations \cite{han2018co}, thereby training networks with improved generalization capabilities. Our distinction from previous studies lies in our investigation of disagreement of unstructured 3D Gaussian radiance fields instead of generalized neural networks, where we consider Gaussian positions and rendered results.

\section{Point Disagreement and Rendering Disagreement}
\label{sec:empirical study}

\subsection{Preliminary and Definition}

\noindent\textbf{\textit{3D Gaussian Splatting.}} 3D Gaussian splatting \cite{kerbl20233d} represents a scene with a set of 3D Gaussians. The $i$-th Gaussian primitive can be described as $\theta_i = \{\mu_i, s_i, q_i, \alpha_i, f_i\}$, where $\mu_i \in \mathbb{R}^3$ is the center, $s_i \in \mathbb{R}^3$ is the scaling factor, $q_i \in \mathbb{R}^4$ is the rotation, $\alpha_i \in \mathbb{R}$ is the opacity for rendering and $f_i \in \mathbb{R}^K$ is the $K$-dimensional color feature. In the three-dimensional space, the influence of the $i$-th Gaussian primitive on the position $x$ is defined as $G_i(x)$ :
\begin{equation}
    G_i(x) = e^{-\frac{1}{2}(x-\mu_i)^T\Sigma_i^{-1}(x-\mu_i)},
\end{equation}
where the covariance matrix $\Sigma$  is calculated from the scale $s$ and rotation $q$. For color rendering, 3D Gaussian Splatting orders all the Gaussians that contribute to each 2D-plane pixel and renders through alpha-blending. For optimization, 3DGS optimizes the parameters $\theta$ for Gaussians through gradient descent and conducts densification to identify and duplicate active Gaussians. 3DGS \cite{kerbl20233d} suggests utilizing the point cloud from COLMAP \cite{schoenberger2016mvs, schoenberger2016sfm} or other SfMs for initialization. However, the number of initial points is limited under sparse views, thus we follow FSGS \cite{zhu2023FSGS} to conduct a stereo-fusion-based initialization.

\noindent\textbf{\textit{Point disagreement.}} When considering the 3-axis position of Gaussians, Gaussians in two radiance fields can be regarded as two point clouds. We evaluate their differences using Fitness and root mean square error (RMSE), which are used to evaluate the registration between point clouds \cite{zhou2018open3d}. Fitness calculates the overlapping area between Gaussians concerning a max distance $\tau=5$ for correspondence. RMSE computes the average distance of correspondent points.

\noindent\textbf{\textit{Rendering disagreement.}} We evaluate the differences by considering both rendered images and depth maps. We use PSNR to compute differences between two rendered images. For depth maps, we use Relative Absolute Error (absErrorRel), which is commonly used to evaluate the accuracy of depth values \cite{Uhrig2017THREEDV}.

\noindent\textbf{\textit{Evaluation for reconstruction quality.}} The most popular method to evaluate the reconstruction is to compare rendered images with ground-truth images at test views. We follow this evaluation and use the PSNR metric. To provide a more comprehensive evaluation, we obtain the ground-truth Gaussian positions and depth maps by training a 3D Gaussian radiance field with dense views for each scene. We use Fitness, RMSE, and absErrorRel for evaluation.

\begin{figure}[t]
    \centering
    \includegraphics[width=1.0\linewidth]{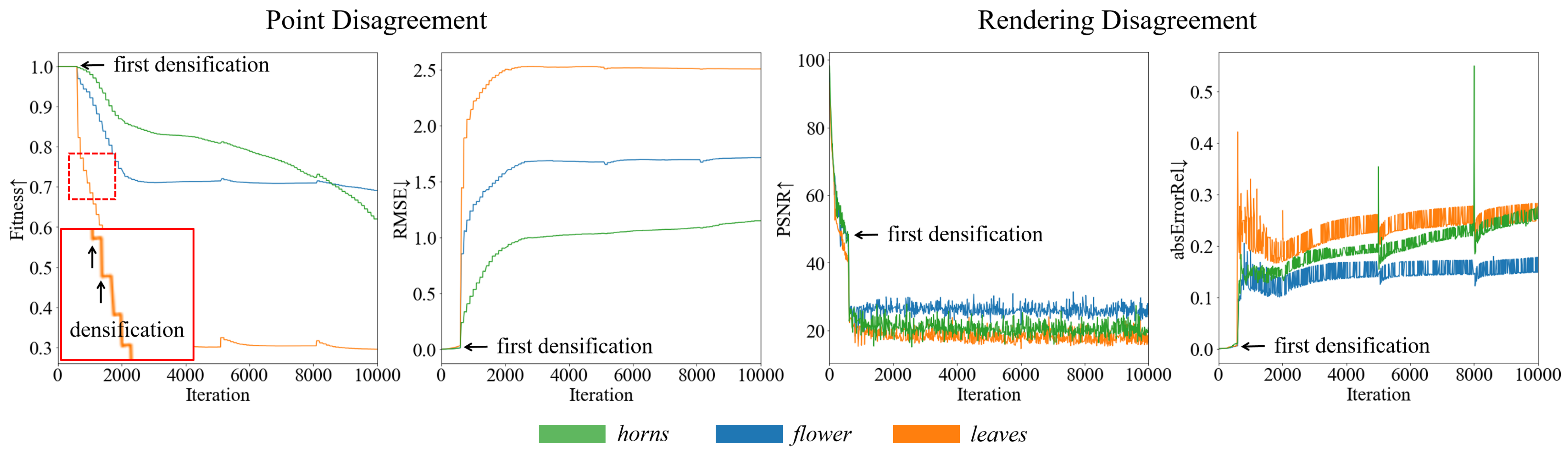}
    \caption{The recorded different behaviors of two 3d Gaussian radiance fields during training. The point disagreement and rendering disagreement increases during training, especially during densification.}
    \label{fig:disagreement during densification}
\end{figure}

\subsection{Empirical Study}

\begin{figure}[t]
    \centering
    \includegraphics[width=1.0\linewidth]{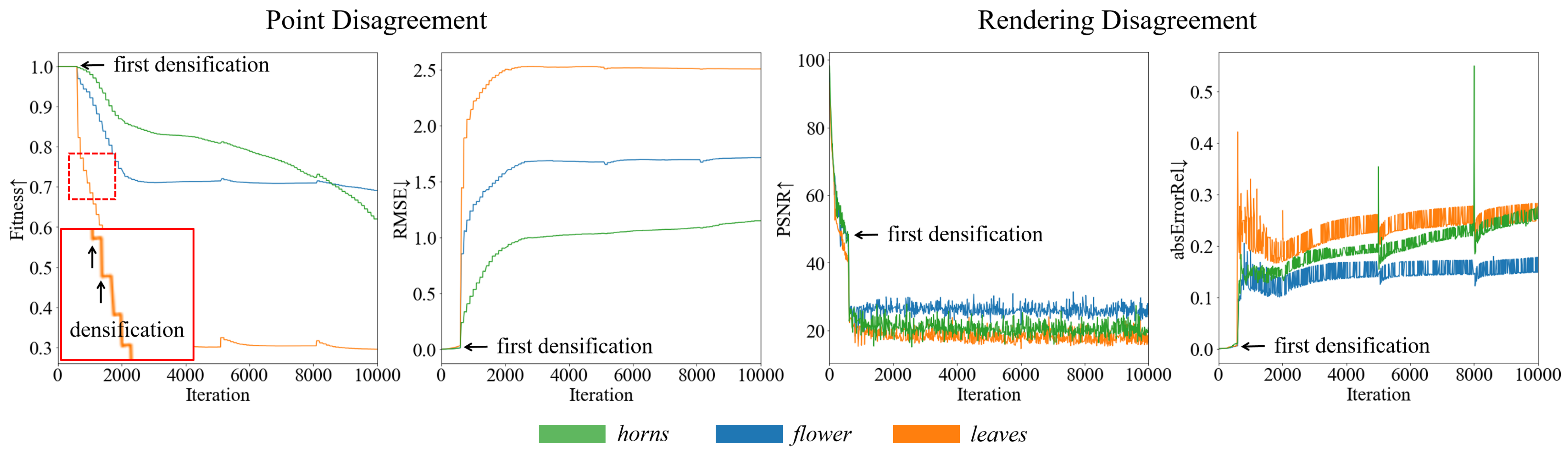}
    \caption{The recorded different behaviors of two 3d Gaussian radiance fields during training. The point disagreement and rendering disagreement increases during training, especially during densification.}
    \label{fig:disagreement during densification}
\end{figure}

\noindent\textbf{\textit{Two 3D Gaussian radiance fields trained with the same sparse views can exhibit different behaviors.}} We simultaneously train two 3D Gaussian radiance fields and record the disagreements during training. Since ground-truth supervision is directly imposed on training views, we evaluate the rendering disagreement at unseen views. As shown in \cref{fig:disagreement during densification}, the two 3D Gaussian radiance fields exhibit different behaviors. In particular, the two disagreements grow significantly during densification, which creates new Gaussians but locates them blindly to the scene geometry. Therefore, the randomness during densification can be a source of error geometry in sparse-view 3DGS.

\begin{figure}[t]
    \centering
    \includegraphics[width=1.0\linewidth]{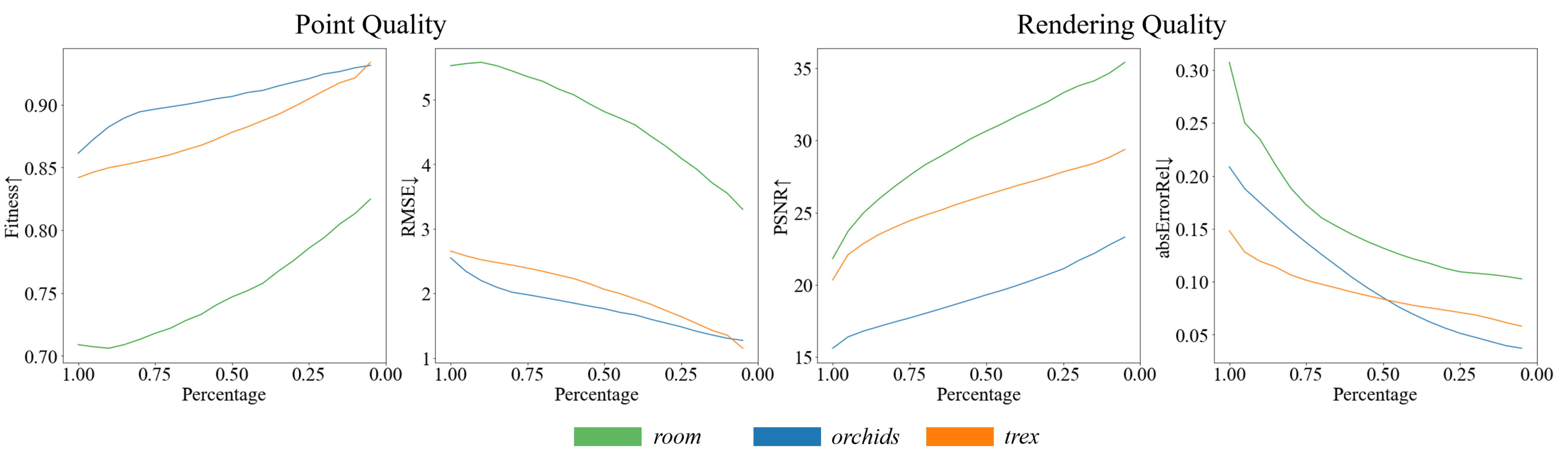}
    \caption{The correlation between the two disagreements and reconstruction quality. The x-axis percentage represents we mask out the percentage of regions with the highest disagreement scores. With the reduction of regions with higher disagreement scores, the reconstruction quality averaging the remaining regions continuously improves.}
    \label{fig:not all equal}
\end{figure}

\begin{figure}[t]
    \centering
    \includegraphics[width=1.0\linewidth]{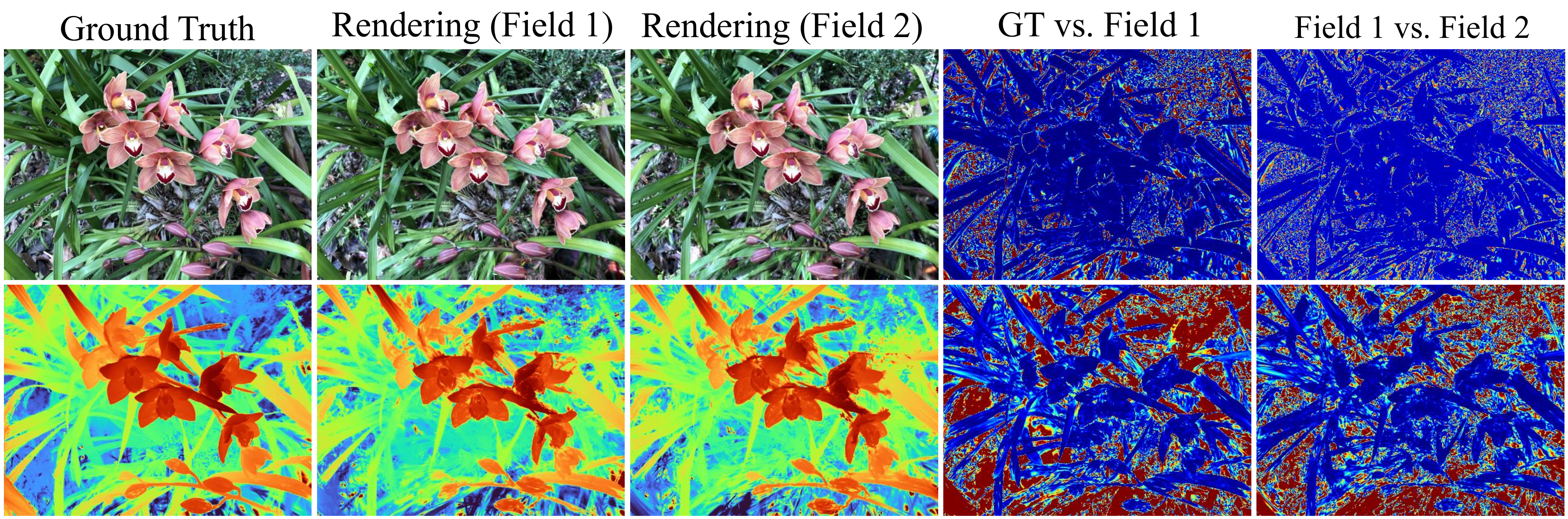}
    \caption{Test-view visualization of rendered images and depth maps, and the corresponding error maps. The disagreed regions between the rendered results of two 3D Gaussian radiance fields tend to be inaccurate compared to ground truth.}
    \label{fig:not all equal vis}
\end{figure}

\noindent\textbf{\textit{Point disagreement and Rendering disagreement are negatively correlated with accurate reconstruction of the scene.}} Based on observing different behaviors between two 3D Gaussian radiance fields, we further investigate whether their disagreements imply certain correlations concerning the reconstructed geometry quality. We mask out areas with a certain percentage with the highest disagreement scores across two 3D Gaussian radiance fields and calculate the reconstruction quality of the remaining regions. As shown in \cref{fig:not all equal}, with the reduction of regions with higher disagreement scores, the reconstruction quality of the remaining regions continuously improves. This demonstrates a negative correlation between the disagreement and accurate reconstruction. Therefore, we can identify the inaccurate reconstruction based on the disagreements even without ground-truth information. We provide the test-view visualization of rendered results with the corresponding error maps in \cref{fig:not all equal vis}. We can observe that the disagreed regions between the rendered results of two 3D Gaussian radiance fields tend to be inaccurate compared to ground truth.

\section{Method}
\label{sec:method}

CoR-GS identifies and suppresses the inaccurate reconstruction based on both point disagreement and rendering disagreement. We simultaneously train two 3D Gaussian radiance fields $\Theta^1  = \{\theta_i^1|i=1,2,...N^1\}$ and $\Theta^2  = \{\theta_i^2|i=1,2,...N^2\}$, where $N^1$ and $N^2$ are the number of Gaussians. After training, we keep one 3D Gaussian radiance filed for inference. In the following content, We omit the process for the second field $\Theta^2$, which is in the same manner for $\Theta^1$.

\subsection{Co-pruning}



The sampling implementation in densification creates new Gaussians blind to the geometry and can be hard to correct with sparse views. Based on the point disagreement, co-pruning identifies and prunes Gaussians that are located in inaccurate positions. We first find the matching correspondence $f: \Theta^1 \to \Theta^2$ based on their nearest Gaussians in the other set: 
\begin{equation}
    f(\theta^1_i) = {\rm KNN}(\theta^1_i, \Theta^2).
\end{equation}
We set a permissible maximum distance $\tau=5$ to calculate the non-matching mask $M$:
\begin{equation}
M_{i} = \left \{
\begin{array}{ll}
1, & \sqrt{(\theta^1_{xi} - f(\theta^1_i)_{x})^2 + (\theta^1_{yi} - f(\theta^1_i)_{y})^2 + (\theta^1_{zi} - f(\theta^1_i)_{z})^2} > \tau, \\
0, & {\rm otherwise}, \\
\end{array}
\right.
\end{equation}
where $\theta^1_{xi}, \theta^1_{yi}, \theta^1_{zi}$ represent the 3-axis positions of Gaussians. After the calculation of the non-matching mask for both sets of Gaussians, we prune all Gaussians that are marked as non-matching, which are considered in inaccurate positions. We perform co-pruning every certain number (we set it to 5) of the optimization/density control interleaves, as we aim to prune Gassuains that optimization cannot handle.

\begin{figure}[tb]
    \centering
    \includegraphics[width=1.0\linewidth]{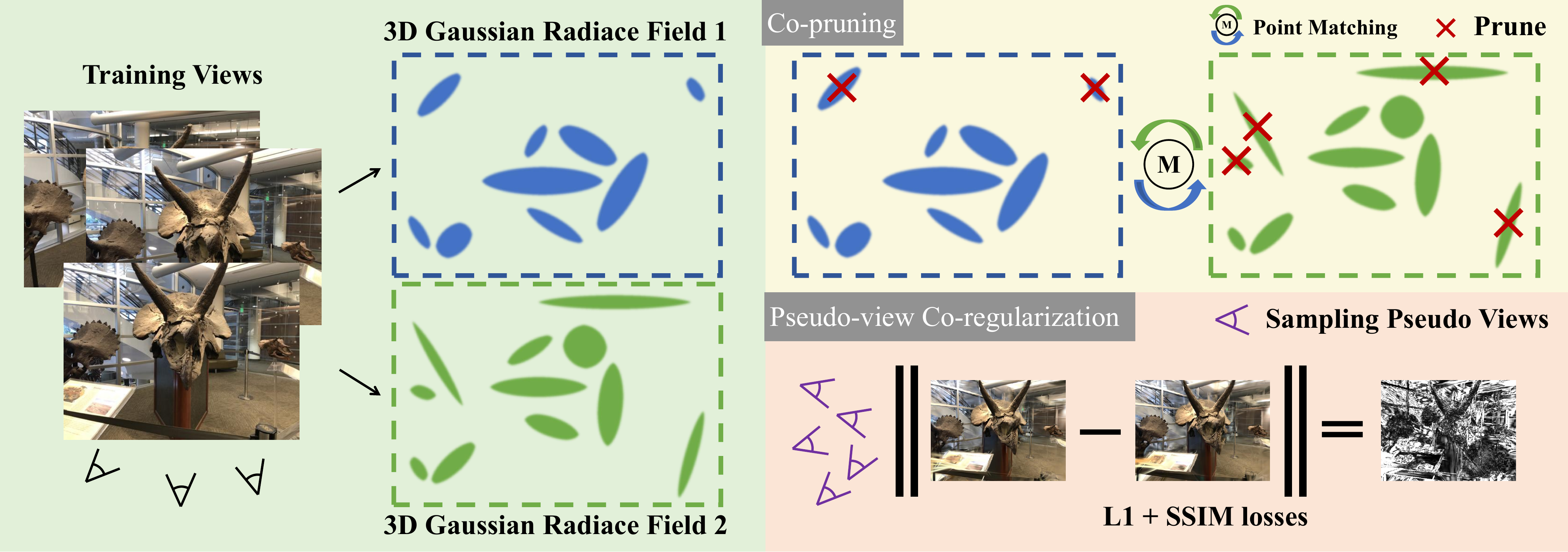}
    \caption{Overview of CoR-GS. We train two 3D Gaussian radiance fields simultaneously and regularize them by suppressing point disagreement and rendering disagreement.}
    \label{fig:method overview}
\end{figure}

\subsection{Pseudo-view Co-regularization}


\noindent\textbf{\textit{Sampling pseudo views.}} The online pseudo view is sampled from the two nearest training views in Euclidean space, following previous studies \cite{zhu2023FSGS}.
\begin{equation}
P' = (t + \epsilon, q).
\end{equation}
Here, $t \in P$ denotes the camera location of training views, $\epsilon$ is the random noise sampled from a normal distribution and $q$ is a quaternion representing the rotation averaged from the two training cameras.

\noindent\textbf{\textit{Color co-regularization.}} At the sampled pseudo view, we render two images $I'^1$ and $I'^2$ using each of the two Gaussian primitive sets $\Theta^1$ and $\Theta^2$, respectively. The color reconstruction loss is a combination of L1 reconstruction loss and a D-SSIM term with the balance weight $\lambda = 0.2$:
\begin{equation}
    \mathcal{R}_{pcolor} = (1 - \lambda) \mathcal{L}_1(I'^1, I'^2) + \lambda \mathcal{L}_{\mathrm{D-SSIM}}(I'^1, I'^2).
\end{equation}
At the training view, we render the image $I^1$ from the primitive set $\Theta^1$ and supervise it with ground truth $I^*$:
\begin{equation}
    \mathcal{L}_{color} = (1 - \lambda) \mathcal{L}_1(I^1, I^*) + \lambda \mathcal{L}_{\mathrm{D-SSIM}}(I^1, I^*).
\end{equation}
The final training loss is a combination of the ground truth supervised loss at the training view and the color co-regularization term at the pseudo view:
\begin{equation}
    \mathcal{L} = \mathcal{L}_{color} + \lambda_p \mathcal{R}_{pcolor},
\end{equation}
where $\lambda_p$ is the balance weight and we set it to 1.0.



\section{Experiments}
\label{sec:experiment}

\subsection{Setup}

\noindent\textbf{\textit{Datasets.}}
We conducted experiments on four datasets: LLFF \cite{mildenhall2019local}, Mip-NeRF360 \cite{barron2022mip360}, DTU \cite{jensen2014large}, and Blender \cite{mildenhall2021nerf}. Our experimental setup aligns with previous studies \cite{niemeyer2022regnerf,yang2023freenerf,wang2023sparsenerf,zhu2023FSGS}, adopting the same data split, with downsampling rates of $8$, $4$, $4$, and $2$ for LLFF, Mip-NeRF360, DTU, and Blender, respectively. More details are in the Materia. To focus on the target object and eliminate background noise during evaluation, we apply object masks similar to prior works \cite{niemeyer2022regnerf} for DTU. Camera poses are assumed to be known based on calibration or other established methods, following the conventions of sparse-view settings.

\noindent\textbf{\textit{Comparison Baselines.}}
We take current SOTA methods Mip-NeRF \cite{barron2021mip}, DietNeRF \cite{jain2021putting}, RegNeRF \cite{niemeyer2022regnerf}, FreeNeRF \cite{yang2023freenerf}, SparseNeRF \cite{wang2023sparsenerf}, and FSGS \cite{zhu2023FSGS} for comparisons. For most methods, we directly report their best published quantitative results. For vanilla 3DGS, we report our implemented results.

\noindent\textbf{\textit{Evaluation Metrics.}}
We present quantitative evaluations of the reconstruction performance through reporting PSNR, SSIM \cite{wang2004ssim}, and LPIPS \cite{zhang2018lpips} scores. Additionally, we calculate an Average Error (AVGE) \cite{niemeyer2022regnerf}, derived from the geometric mean of $\text{MSE} = 10^{-\text{PSNR}/10}$, $\sqrt{1 - \text{SSIM}}$, and LPIPS.

\noindent\textbf{\textit{Implementations.}}
We conduct training for 10k iterations on the LLFF, DTU, and Blender datasets and follow 3DGS \cite{kerbl20233d} to train models for 30k iterations on the MipNeRF dataset. Following FSGS \cite{zhu2023FSGS}, we initialize 3DGS and CoR-GS with fused stereo point cloud from sparse views.

\subsection{Comparison}

\begin{table}[!t]   
  \caption{Quantitative results on LLFF with 3, 6, 9 training views. The best, second-best, and third-best entries are marked in red, orange, and yellow, respectively.}
  \label{tab:llff}
  \centering
  \resizebox{1\linewidth}{!}{
  \begin{tabular}{@{}l|ccc|ccc|ccc|ccc}
    \toprule
     \multirow{2}{*}{Method}   & \multicolumn{3}{c|}{PSNR$\uparrow$}  & \multicolumn{3}{c|}{SSIM$\uparrow$}   & \multicolumn{3}{c|}{LPIPS$\downarrow$}  & \multicolumn{3}{c}{AVGE$\downarrow$} \\
      &  3-view & 6-view & 9-view & 3-view & 6-view & 9-view & 3-view & 6-view & 9-view & 3-view & 6-view & 9-view \\
    \midrule
    Mip-NeRF \cite{barron2021mip}   & 16.11 & 22.91 & 24.88 & 0.401 & 0.756 & 0.826 & 0.460 & 0.213 & 0.160 & 0.215 & 0.090 & 0.066  \\
    DietNeRF \cite{jain2021putting} & 14.94 & 21.75 & 24.28 & 0.370 & 0.717 & 0.801 & 0.496 & 0.248 & 0.183 & 0.240 & 0.105 & 0.073 \\
    RegNeRF \cite{niemeyer2022regnerf}  & 19.08 & 23.10 & 24.86 & 0.587 & 0.760 & 0.820 & 0.336 & 0.206 & 0.161 & 0.149 & 0.086 & 0.067 \\
    FreeNeRF \cite{yang2023freenerf}   & 19.63 & 23.73 & 25.13 & 0.612 & 0.779 & 0.827 & 0.308 & 0.195 & 0.160 & 0.134 & 0.075 & 0.064 \\
    SparseNeRF \cite{wang2023sparsenerf}   & \cellcolor[HTML]{FFFFD4}19.86 & - & - & 0.624 & - & - & 0.328 & - & - & 0.127 &- & - \\
    3DGS \cite{kerbl20233d}  & 19.22 & \cellcolor[HTML]{FFFFD4}23.80 & \cellcolor[HTML]{FFE4CF}25.44 &  \cellcolor[HTML]{FFFFD4}0.649 & \cellcolor[HTML]{FFFFD4}0.814 & \cellcolor[HTML]{FFE4CF}0.860 & \cellcolor[HTML]{FFE4CF}0.229 & \cellcolor[HTML]{FFE4CF}0.125 & \cellcolor[HTML]{FFE4CF}0.096 & \cellcolor[HTML]{FFFFD4}0.120 & \cellcolor[HTML]{FFE4CF}0.066 & \cellcolor[HTML]{FFE4CF}0.051 \\
    FSGS \cite{zhu2023FSGS} & \cellcolor[HTML]{FFE4CF}20.43 & \cellcolor[HTML]{FFE4CF}24.09 & \cellcolor[HTML]{FFFFD4}25.31 & \cellcolor[HTML]{FFE4CF}0.682 & \cellcolor[HTML]{FFE4CF}0.823 & \cellcolor[HTML]{FFE4CF}0.860 & \cellcolor[HTML]{FFFFD4}0.248 & \cellcolor[HTML]{FFFFD4}0.145 & \cellcolor[HTML]{FFFFD4}0.122 & \cellcolor[HTML]{FFE4CF}0.104 & \cellcolor[HTML]{FFE4CF}0.066 & \cellcolor[HTML]{FFFFD4}0.054 \\
    CoR-GS (ours) & \cellcolor[HTML]{FFCCC9}20.45 & \cellcolor[HTML]{FFCCC9}24.49 & \cellcolor[HTML]{FFCCC9}26.06 & \cellcolor[HTML]{FFCCC9}0.712 & \cellcolor[HTML]{FFCCC9}0.837 & \cellcolor[HTML]{FFCCC9}0.874 & \cellcolor[HTML]{FFCCC9}0.196 & \cellcolor[HTML]{FFCCC9}0.115 & \cellcolor[HTML]{FFCCC9}0.089 & \cellcolor[HTML]{FFCCC9}0.101 & \cellcolor[HTML]{FFCCC9}0.060 & \cellcolor[HTML]{FFCCC9}0.046 \\
  \bottomrule
  \end{tabular}
  }
\end{table}

\begin{figure}[!t]
    \centering
    \includegraphics[width=1.0\linewidth]{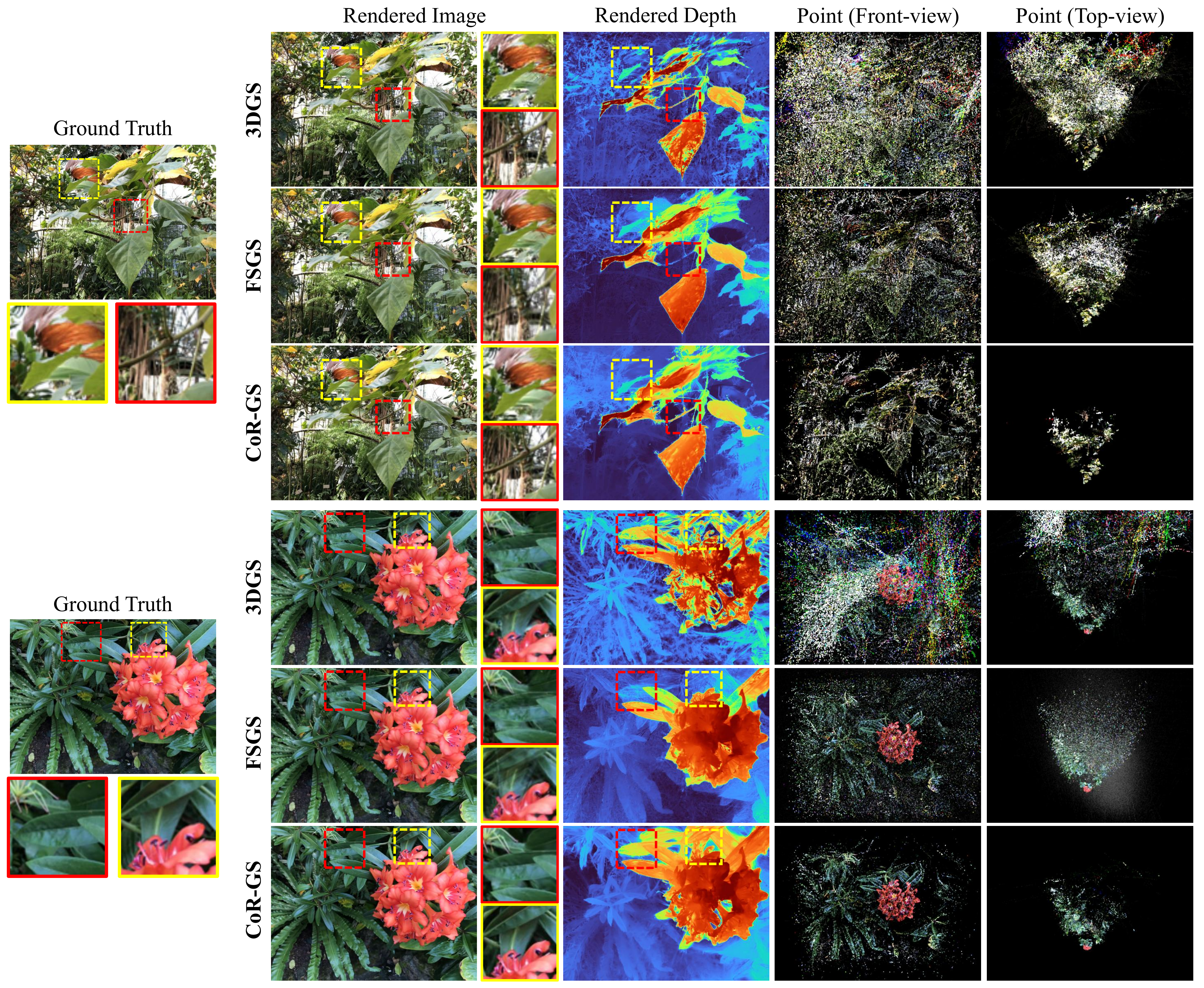}  
    \caption{Qualitative comparison on LLFF. We provide visualizations of Gaussian primitive positions from both the frontal and top views. 3DGS \cite{kerbl20233d} reconstructed incorrect geometry, reflected in the rendered depth maps and the positions of 3D Gaussians. FSGS \cite{zhu2023FSGS} is affected by noises in the external depth and also struggles to reconstruct compact Gaussians. Conducting co-regularization between different primitive sets, our CoR-GS reconstructs the compact scene geometry.
    }
    \label{fig:llff}
\end{figure}

\noindent\textbf{\textit{LLFF.}} We provide the quantitative results on the LLFF dataset in \cref{tab:llff}. Our method consistently achieves the best performance across PSNR, SSIM, LPIPS, and AVGE metrics with 3, 6, and 9 training views. The depth-supervised FSGS \cite{zhu2023FSGS} enhances the performance of 3DGS in scenarios with 3 and 6 views. However, with 9 training views, 3DGS can already reconstruct the scene structure well, which limits the effectiveness of external depth supervision. Our method conducts co-regularization to avoid reconstructing incorrect geometry, leading to improvements across all metrics with various training views. We provide quantitative visualizations of rendered images and depth at novel views and Gaussian points in \cref{fig:llff}. From the rendered depth maps, we observe that vanilla 3DGS can recover certain scene structures but exhibits geometric errors, particularly evident in the unrealistic parts of rendered images. For the point representations, many Gaussians are dispersed throughout space and far from the reconstructed scene. These Gaussians, when observed from novel views, can result in unrealistic rendering images. With depth supervision, FSGS can correct the incorrect geometry of 3DGS; however, noises in external depth maps also adversely affect the geometry. Similarly, relying on depth supervision makes it challenging to directly constrain Gaussians. It can be seen that the Gaussians' dispersion of FSGS is still not compact. Our method effectively assists Gaussians in reconstructing coherent and compact geometry, resulting in higher-quality novel view rendering.

\begin{table}[!t]
  \caption{Quantitative results on Mip-NeRF360. The best are marked in \textbf{bold}.}
  \label{tab:mipnerf360}
  \centering
  \resizebox{1.0\linewidth}{!}{
  \setlength{\tabcolsep}{2mm}
  \begin{tabular}{@{}l|cccc|cccc}
    \toprule
    \multirow{2}{*}{Method} & \multicolumn{4}{c|}{12-view} & \multicolumn{4}{c}{24-view}  \\
    & PSNR $\uparrow$ & SSIM $\uparrow$  & LPIPS $\downarrow$ & AVGE$\downarrow$ & PSNR $\uparrow$ & SSIM $\uparrow$  & LPIPS $\downarrow$ & AVGE$\downarrow$  \\
    \midrule
    3DGS \cite{kerbl20233d} & 18.52 &  0.523 &  \textbf{0.415} & 0.167 &  22.80 &  0.708 &  0.276 & 0.096 \\
    FSGS \cite{zhu2023FSGS} &  18.80 &  0.531 &  0.418 & 0.163 &  23.28 &  0.715 &  0.274 & 0.093 \\
    CoR-GS (ours) & \textbf{19.52}  & \textbf{0.558}  & 0.418 & \textbf{0.151}   &  \textbf{23.39} &  \textbf{0.727} &  \textbf{0.271} & \textbf{0.091} \\
  \bottomrule
  \end{tabular}
  }
\end{table}

\begin{figure}[!t]
    \centering
    \includegraphics[width=0.90\linewidth]{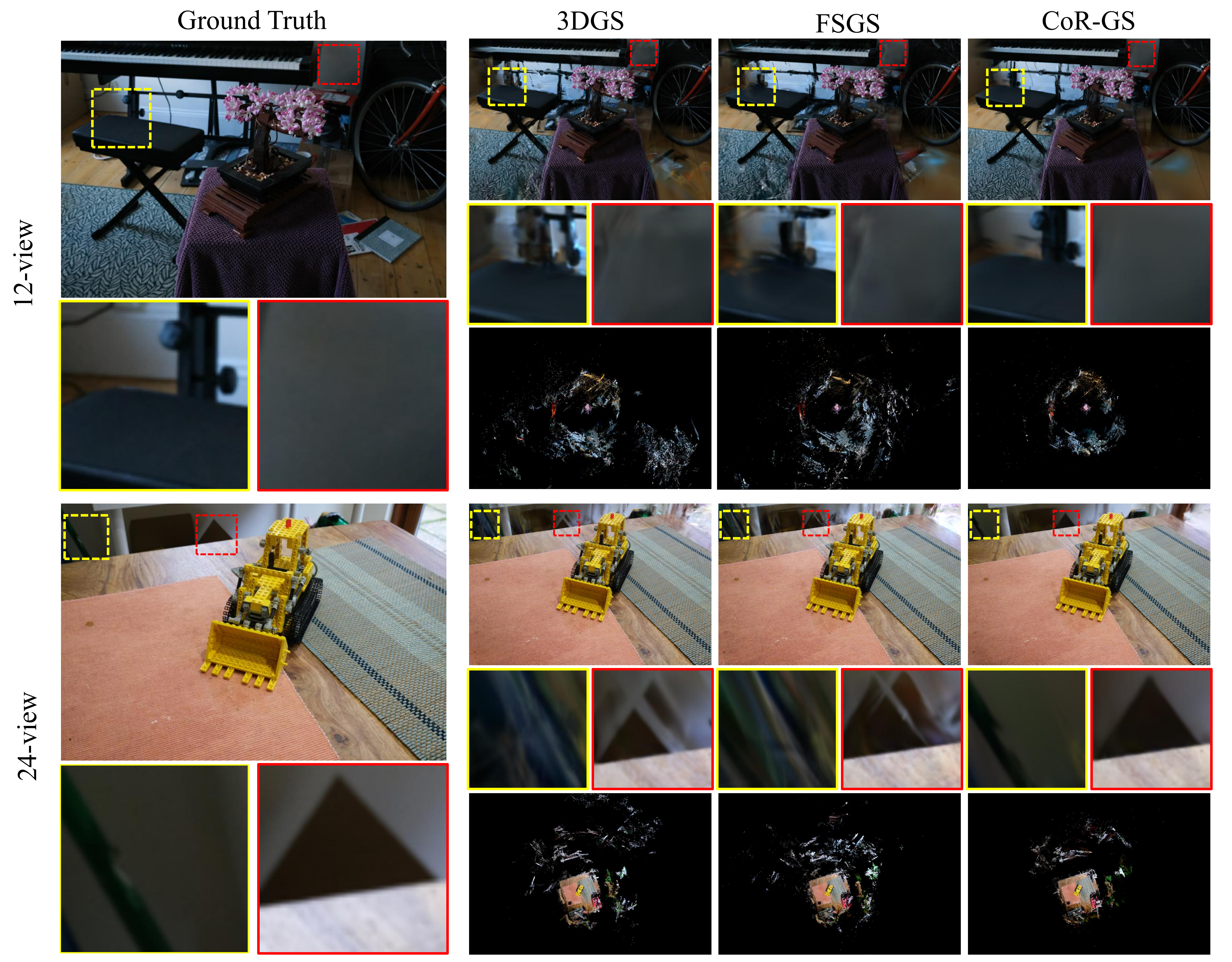}
    \caption{Qualitative comparison on Mip-NeRF360. All three 3DGS-based methods perform well in reconstructing central objects. Our method demonstrates an advantage in non-central regions without sufficient co-observation.}
    \label{fig:mipnerf360}
\end{figure}

\noindent\textbf{\textit{Mip-NeRF360.}} The quantitative results on the Mip-NeRF360 dataset with sparse training views are reported in Table \ref{tab:mipnerf360}. Our method achieves the best performance across PSNR, SSIM, and AVGE with 12 and 24 training views. Vanilla 3DGS achieves the best LPIPS scores with 12 training views. This is because for 360-degree panoramic scenes, 12 views cause many regions to lack co-visibility constraints. In these areas, the pseudo-view co-regularization tends to produce a smoother effect. When using 24 training views, where co-visible regions increase, our method achieves the best LPIPS score. We provide qualitative visualizations in \cref{fig:mipnerf360}. From rendered images, we observe that all methods can achieve good reconstruction for central objects observed from multiple training views. The advantage of our method is evident in its ability to reconstruct clearer structures in non-central regions. From Gaussian positions, we observe that despite training with 360-degree views, 3DGS and FSGS still reconstruct dispersed Gaussians in the distance. The dispersed Gaussians have a negative impact when observing non-central areas in novel views. In contrast, our method reconstructs more compact representations. The results demonstrate that CoR-GS remains well-suited for reconstructing full 360° unbounded scenes with sparse training views.

\begin{table}[tb]
  \caption{Quantitative results on DTU with 3, 6, 9 training views. The best, second-best, and third-best entries are marked in red, orange, and yellow, respectively.}
  \label{tab:dtu}
  \centering
  \resizebox{1\linewidth}{!}{
  \begin{tabular}{@{}l|ccc|ccc|ccc|ccc}
    \toprule
     \multirow{2}{*}{Method}     & \multicolumn{3}{c|}{PSNR$\uparrow$}  & \multicolumn{3}{c|}{SSIM$\uparrow$}   & \multicolumn{3}{c|}{LPIPS$\downarrow$}  & \multicolumn{3}{c}{AVGE$\downarrow$} \\
      &   3-view & 6-view & 9-view & 3-view & 6-view & 9-view & 3-view & 6-view & 9-view & 3-view & 6-view & 9-view \\
    \midrule
    DietNeRF \cite{jain2021putting}  & 11.85 & 20.63 & 23.83 & 0.633 & 0.778 & 0.823 & 0.314 & 0.201 & 0.173 & 0.243 & 0.101 & 0.068 \\
    RegNeRF \cite{niemeyer2022regnerf}   & 18.89 & 22.20 & 24.93 & 0.745 & 0.841 & 0.884 & 0.190 & 0.117 & 0.089 & 0.112 & 0.071 & 0.047 \\
    Mip-NeRF \cite{barron2021mip}    & 9.10 & 16.84 & 23.56 & 0.578 & 0.754 & 0.877 & 0.348 & 0.197 & 0.100 & 0.311 & 0.144 & 0.057  \\
    FreeNeRF \cite{yang2023freenerf}    & \cellcolor[HTML]{FFCCC9}19.92 & \cellcolor[HTML]{FFFFD4}23.25 & \cellcolor[HTML]{FFFFD4}25.38 & \cellcolor[HTML]{FFFFD4}0.787 & \cellcolor[HTML]{FFFFD4}0.844 & \cellcolor[HTML]{FFFFD4}0.888 & \cellcolor[HTML]{FFFFD4}0.182 & \cellcolor[HTML]{FFFFD4}0.137 & \cellcolor[HTML]{FFFFD4}0.096 & \cellcolor[HTML]{FFE4CF}0.098 & \cellcolor[HTML]{FFFFD4}0.068 & \cellcolor[HTML]{FFFFD4}0.046 \\
    SparseNeRF \cite{wang2023sparsenerf}   & \cellcolor[HTML]{FFE4CF}19.55 & - & - & 0.769 & - & - & 0.201 & - & -  & \cellcolor[HTML]{FFFFD4}0.102 & - & - \\
    3DGS \cite{kerbl20233d}  & 17.65 & \cellcolor[HTML]{FFE4CF}24.00 & \cellcolor[HTML]{FFE4CF}26.85 & \cellcolor[HTML]{FFE4CF}0.816 & \cellcolor[HTML]{FFE4CF}0.907 & \cellcolor[HTML]{FFE4CF}0.942 & \cellcolor[HTML]{FFE4CF}0.146 & \cellcolor[HTML]{FFE4CF}0.076 & \cellcolor[HTML]{FFE4CF}0.049 & 0.108 & \cellcolor[HTML]{FFE4CF}0.050 & \cellcolor[HTML]{FFE4CF}0.032 \\
    CoR-GS (ours)  & \cellcolor[HTML]{FFFFD4}19.21 & \cellcolor[HTML]{FFCCC9}24.51 & \cellcolor[HTML]{FFCCC9}27.18 & \cellcolor[HTML]{FFCCC9}0.853 & \cellcolor[HTML]{FFCCC9}0.917 & \cellcolor[HTML]{FFCCC9}0.947 & \cellcolor[HTML]{FFCCC9}0.119 & \cellcolor[HTML]{FFCCC9}0.068 & \cellcolor[HTML]{FFCCC9}0.045 & \cellcolor[HTML]{FFCCC9}0.087 & \cellcolor[HTML]{FFCCC9}0.046 & \cellcolor[HTML]{FFCCC9}0.029 \\
  \bottomrule
  \end{tabular}
      }
\end{table}

\begin{figure}[!t]
    \centering
    \includegraphics[width=1.0\linewidth]{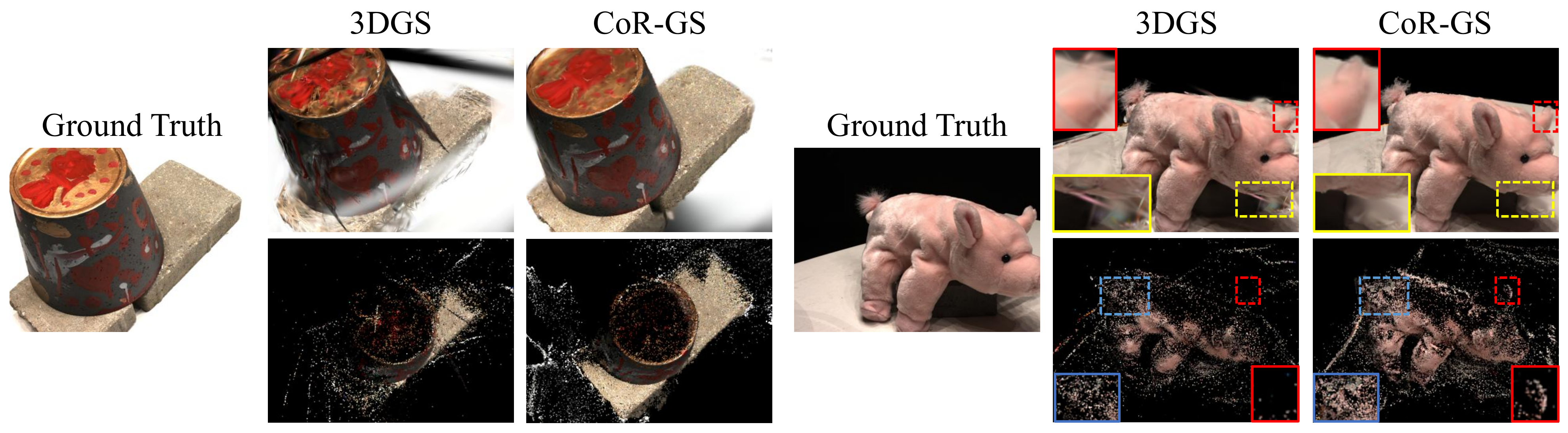}
    \caption{Qualitative results on DTU. CoR-GS reconstructs more complete objects.}
    \label{fig:dtu}
\end{figure}

\noindent\textbf{\textit{DTU.}} The quantitative results on the DTU dataset are reported in Table \ref{tab:dtu}. Our method achieves the best in SSIM, LPIPS, and AVGE with 3, 6, and 9 training views. However, with only 3 views, we observe that 3DGS-based methods get lower PSNR scores compared to NeRF-based methods. This is caused by that 3DGS simply renders the black background in invisible areas beyond training views and the 3 training views of DTU cause lots of invisible areas. NeRFs achieve better PSNR scores due to their interpolation nature for invisible areas. With the increase in training views, we see that 3DGS-based methods get better PSNR scores compared to NeRFs. In the qualitative visualizations in Figure \ref{fig:dtu}, our method renders more complete objects and reconstructs more compact Gaussians compared to 3DGS.

\noindent\textbf{\textit{Blender.}} The quantitative results on the Blender dataset with 8 surrounding training views are reported in \cref{tab:blender}. Our method gets the best scores in SSIM and LPIPS, and second in PSNR. The results demonstrate that CoR-GS is also applicable for reconstructing complex objects.

\begin{table}[tb] 
  \caption{Quantitative results on Blender with 8 training views. The best, second-best, and third-best entries are marked in red, orange, and yellow, respectively.}
  \label{tab:blender}
  \centering
  \resizebox{0.8\linewidth}{!}{
  \setlength{\tabcolsep}{9mm}
  \begin{tabular}{@{}l|ccc}
    \toprule
     Method & PSNR $\uparrow$ & SSIM $\uparrow$  & LPIPS $\downarrow$ \\
    \midrule
    RegNeRF \cite{niemeyer2022regnerf} & 23.86 & 0.852 & 0.105 \\
    FreeNeRF \cite{yang2023freenerf} & \cellcolor[HTML]{FFFFD4}24.26 & \cellcolor[HTML]{FFFFD4}0.883 & \cellcolor[HTML]{FFFFD4}0.098 \\
    SparseNeRF \cite{wang2023sparsenerf} & 24.04 & 0.876 & 0.113 \\
    3DGS \cite{kerbl20233d} & 23.20 &  0.870 & 0.104 \\
    FSGS \cite{zhu2023FSGS} & \cellcolor[HTML]{FFCCC9}24.64 & \cellcolor[HTML]{FFE4CF}0.895 &  \cellcolor[HTML]{FFE4CF}0.095 \\
    CoR-GS (ours) & \cellcolor[HTML]{FFE4CF}24.43 & \cellcolor[HTML]{FFCCC9}0.896 & \cellcolor[HTML]{FFCCC9}0.084 \\
  \bottomrule
  \end{tabular}
  }
\end{table}

\begin{table}[!t]
  \caption{Efficiency comparison. Our method achieves the best novel view synthesis and the least inference cost.}
  \label{tab:efficiency}
  \centering
  \resizebox{1.0\linewidth}{!}{
  \setlength{\tabcolsep}{3mm}
  \begin{tabular}{@{}l|ccccc|cc}
    \toprule
    \multirow{2}{*}{Method} & \multicolumn{5}{c|}{Inference} & \multicolumn{2}{c}{Train} \\
      & Points & FPS  & PSNR $\uparrow$ & SSIM $\uparrow$  & LPIPS $\downarrow$ &  Time & GPU Mem  \\
    \midrule
    FreeNeRF \cite{yang2023freenerf} & - & 0.09 & 19.63 & 0.612 & 0.308 & 2.3h & 4$\times$48 GB \\
    SparseNeRF \cite{wang2023sparsenerf} & - & 0.09 & 19.86 & 0.624 & 0.328 & 1.5h & 32GB \\
    3DGS \cite{kerbl20233d} & 1.16$\times$10$^{5}$ & 318 & 19.22 & 0.649 & 0.229 & \textbf{2.5min} & \textbf{2GB} \\
     CoR-GS (ours) & \textbf{7.85$\times$10$^{4}$}  & \textbf{349} & \textbf{20.45} & \textbf{0.712} & \textbf{0.196} & 6min & 3GB \\
  \bottomrule
  \end{tabular}
  }
\end{table}

\noindent\textbf{\textit{Efficiency.}}
In \cref{tab:efficiency}. We conduct an efficiency comparison of both training and inference on the LLFF 3-view setting with an RTX 3090 Ti GPU. Compared to NeRFs, 3DGS has significant advantages of efficiency. Compared to vanilla 3DGS, our method introduces some additional training costs due to training two 3D  Gaussian radiance feilds and rendering pseudo views but remains much more efficient than NeRFs. Our method is the most efficient during inference. Due to reconstructing more compact representations, our method reduces the number of Gaussians by $33\%$ and thus improves inference speed.

\subsection{Ablation Study}

\begin{figure}[!t]
    \centering
    \includegraphics[width=0.9\linewidth]{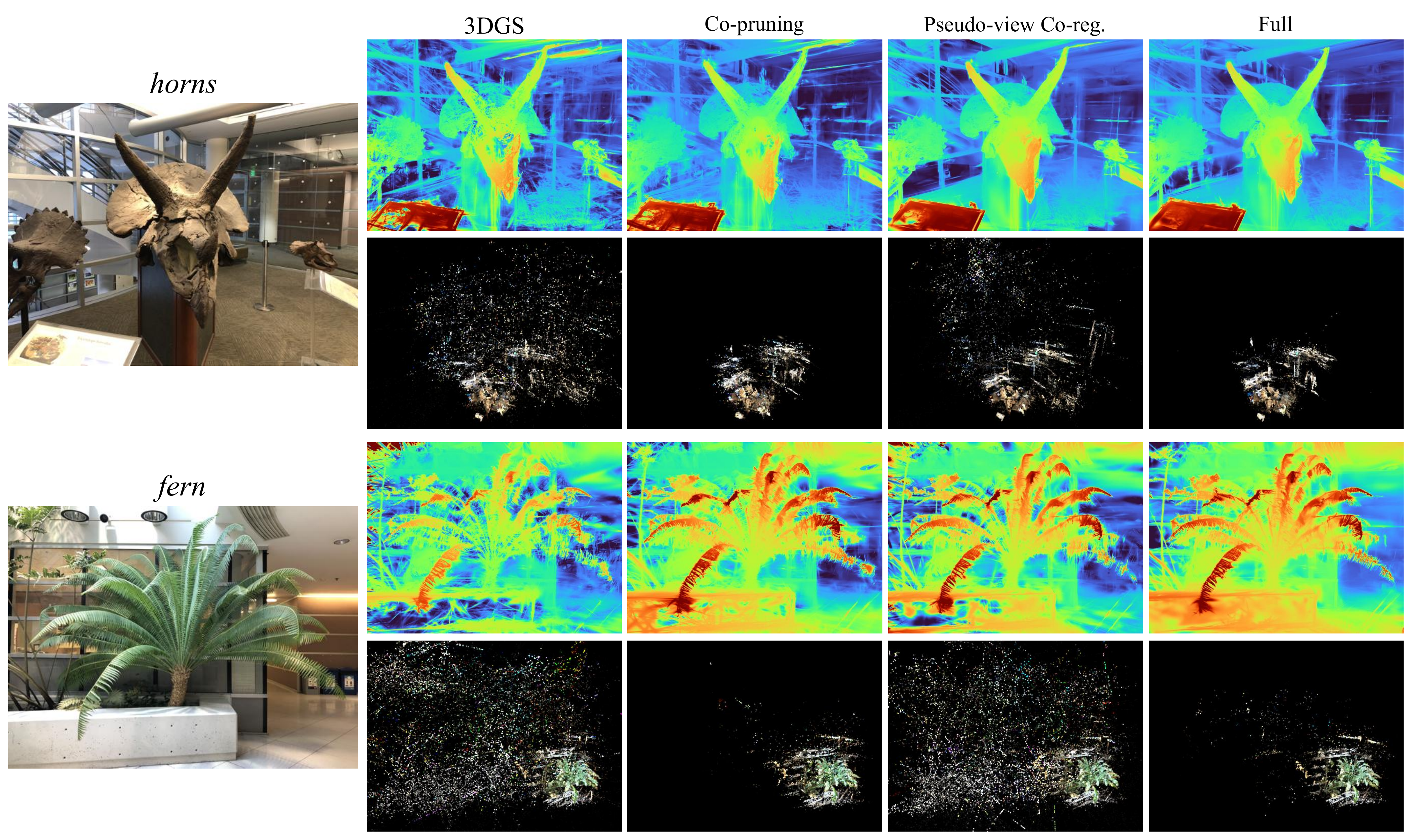}  
    \caption{Visualization results for ablation. Co-pruning prunes Gaussians far from the reconstructed scene for compact representations, while pseudo-view co-regularization helps correct nearby Gaussians, complementing each other in our method.}
    \label{fig:ablation}
\end{figure}

We ablate the effectiveness of suppressing the point disagreement and rendering disagreement. The quantitative results are reported in \cref{tab:ablation}. The results indicate that suppressing each disagreement is individually beneficial for sparse-view 3DGS, and the combination of both achieves the best performance. We further provide the qualitative visualizations in \cref{fig:ablation} to better illustrate their effects.

\noindent\textbf{\textit{Co-pruning.}} Co-pruning improves the geometric structure, reflected in more reasonable depth rendering and more compact Gaussians. However, its drawback is that the rendered depth lacks smoothness, especially evident in the floor part of the horns scene. This is because co-pruning mainly operates on non-matching Gaussians that are far from reconstructed scenes while lacking sufficient constraints on matching Gaussians that are located in reasonable positions.

\noindent\textbf{\textit{Pseudo-view Co-regularization.}} Solely with pseudo-view co-regularization is effective in constraining Gaussians to obtain reasonable depth maps due to inaccurately located Gaussians can be identified through rendering disagreement. However, due to the lack of direct constraints on Gaussians, the representations are not compact enough, leading to some geometric errors that cannot be effectively corrected, especially evident in the fern scene. The full set of CoR-GS compresses two kinds of disagreement, reconstructing more accurate scene geometry and compact representations.

\begin{table}[!t] 
  \caption{Ablation study of CoR-GS with 3 training views.}
  \label{tab:ablation}
  \centering
  \resizebox{1.0\linewidth}{!}{
  \setlength{\tabcolsep}{2mm}
  \begin{tabular}{@{}cc|ccc|ccc}
    \toprule
     \multicolumn{2}{c|}{CoR-GS} & \multicolumn{3}{c|}{LLFF} & \multicolumn{3}{c}{DTU}  \\
     Co-Pruning &  Pseudo-view Co-reg. & PSNR $\uparrow$ & SSIM $\uparrow$  & LPIPS $\downarrow$  & PSNR $\uparrow$ & SSIM $\uparrow$  & LPIPS $\downarrow$  \\
    \midrule
    & & 19.22 & 0.649 & 0.229 & 17.65 & 0.816 & 0.146 \\
    \hline
     \checkmark & & 19.62 & 0.673 & 0.217 & 18.59 & 0.836 & 0.127    \\
      &  \checkmark & 20.26 & 0.706   & 0.198   & 18.56  &  0.849 & 0.128\\
    \checkmark &  \checkmark & \textbf{20.45} & \textbf{0.712} & \textbf{0.196} & \textbf{19.21} & \textbf{0.853} & \textbf{0.119} \\
  \bottomrule
  \end{tabular}
  }
\end{table}

\section{Conclusion}
\label{sec:conclusion}

This paper introduces a new co-regularization perspective for improving sparse-view 3DGS. We observe the two 3D Gaussian radiance fields exhibit different behaviors for the same scene with sparse training views and propose point disagreement and rendering disagreement to quantitatively indicate the differences. We further demonstrate the negative correlation between the two disagreements and accurate reconstruction, which allows us to identify inaccurate reconstruction unsupervisedly. Based on the study, we propose CoR-GS, which improves sparse-view 3DGS by depressing the two kinds of disagreement. We validate the effectiveness of CoR-GS across various datasets, achieving state-of-the-art novel view synthesis in the sparse-view setting.

\title{Supplementary Material for CoR-GS}

\title{Supplementary Material for CoR-GS} 

\titlerunning{CoR-GS}

\author{}

\authorrunning{J. Zhang et al.}


\institute{}

\maketitle

\setcounter{section}{0}
\setcounter{subsection}{0}
\setcounter{figure}{0}
\setcounter{table}{0}

\renewcommand\thesection{\Alph{section}}
\renewcommand\thefigure{\Roman{figure}}
\renewcommand\thetable{\Roman{table}}

\section*{Overview}
We organize the material as follows. We first provide more results of the point disagreement and rendering disagreement in \cref{suppl_sec:disagreement}. Then, the hyperparameters and the design of our method are discussed in \cref{suppl_sec:ablation}. We provide more comparison results in \cref{suppl_sec:more comparison}. We provide more visualization results in \cref{suppl_sec:visualization}. The discussion of futures works is in \cref{suppl_sec:future}. Finally, we provide more details of the experiment setup in \cref{suppl_sec:details}.

\section{Additional Results of Disagreement}
\label{suppl_sec:disagreement}

\begin{figure}[!b]
    \centering
    \includegraphics[width=1.0\linewidth]{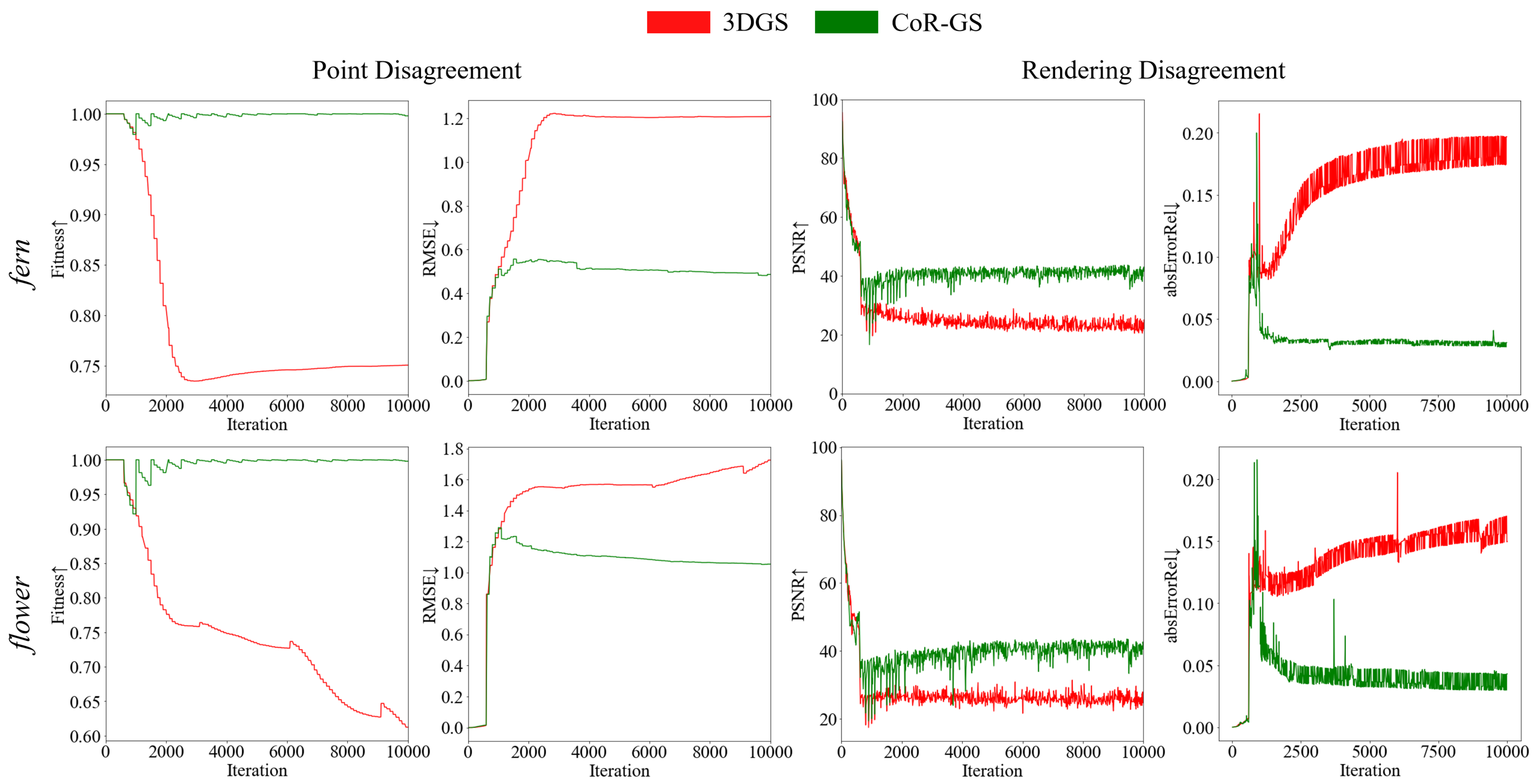}
    \caption{The recorded different behaviors of two 3d Gaussian radiance fields on the 3-view LLFF dataset during training.}
    \label{suppl_fig:disagreement during densification}
\end{figure}

\begin{figure}[!t]
    \centering
    \includegraphics[width=1.0\linewidth]{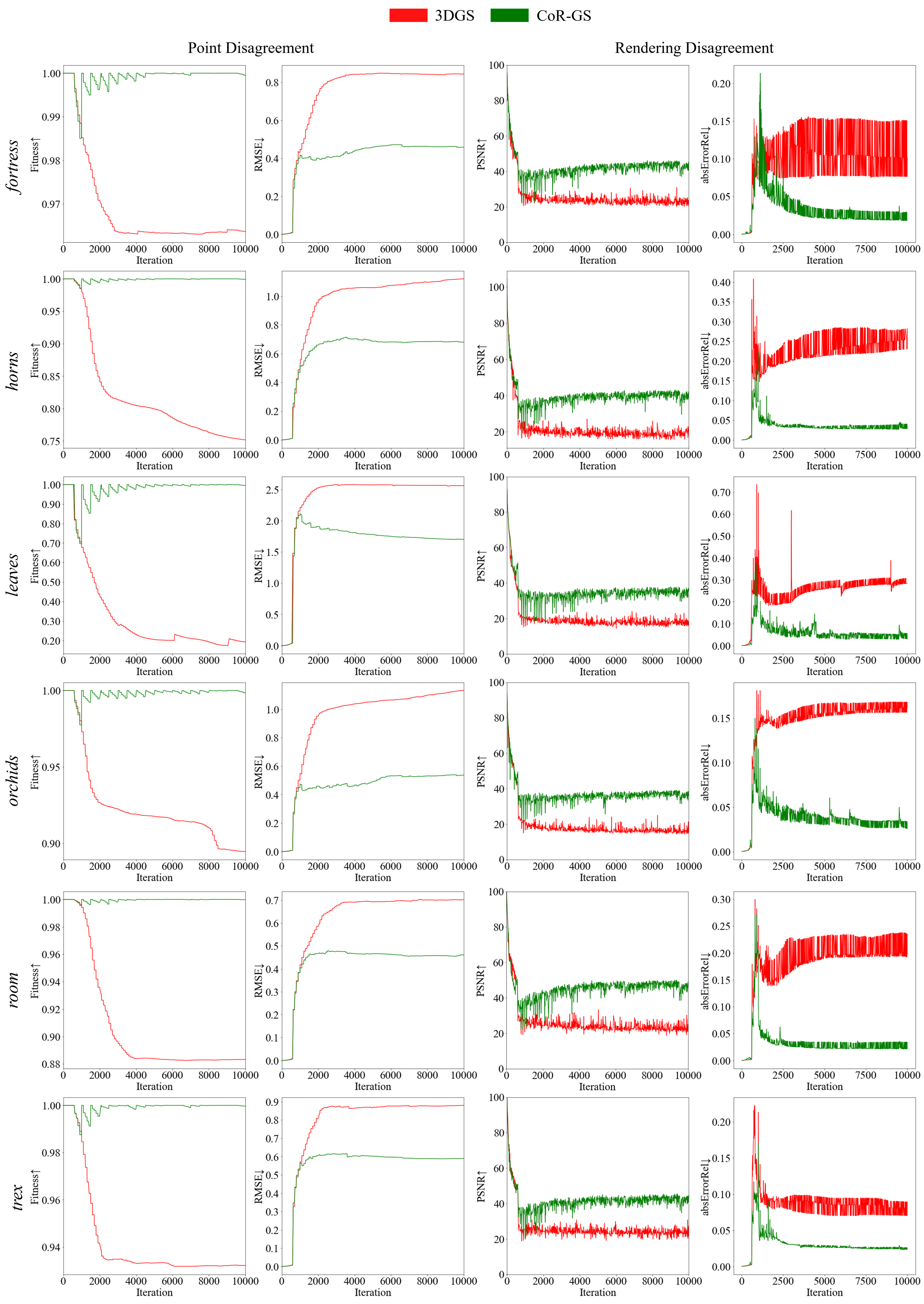}
    \caption{The recorded different behaviors of two 3d Gaussian radiance fields on the 3-view LLFF dataset during training.}
    \label{suppl_fig:disagreement 2 during densification}
\end{figure}

\subsection{Behaviors of Disagreement}

We provide the point disagreement and the rendering disagreement recorded in more scenes. We also provide the two disagreements of our CoR-GS. The results are shown in \cref{suppl_fig:disagreement during densification,suppl_fig:disagreement 2 during densification}. We observe that the two 3D Gaussian radiance fields exhibit different behaviors in various scenes and the disagreements increase significantly during densification. Integrating co-pruning and pseudo-view co-regularization, CoR-GS effectively suppresses the point disagreement and the rendering disagreement of vanilla 3DGS.

\subsection{Disagreement and Accurate Reconstruction}

\begin{figure}[!t]
    \centering
    \includegraphics[width=1.0\linewidth]{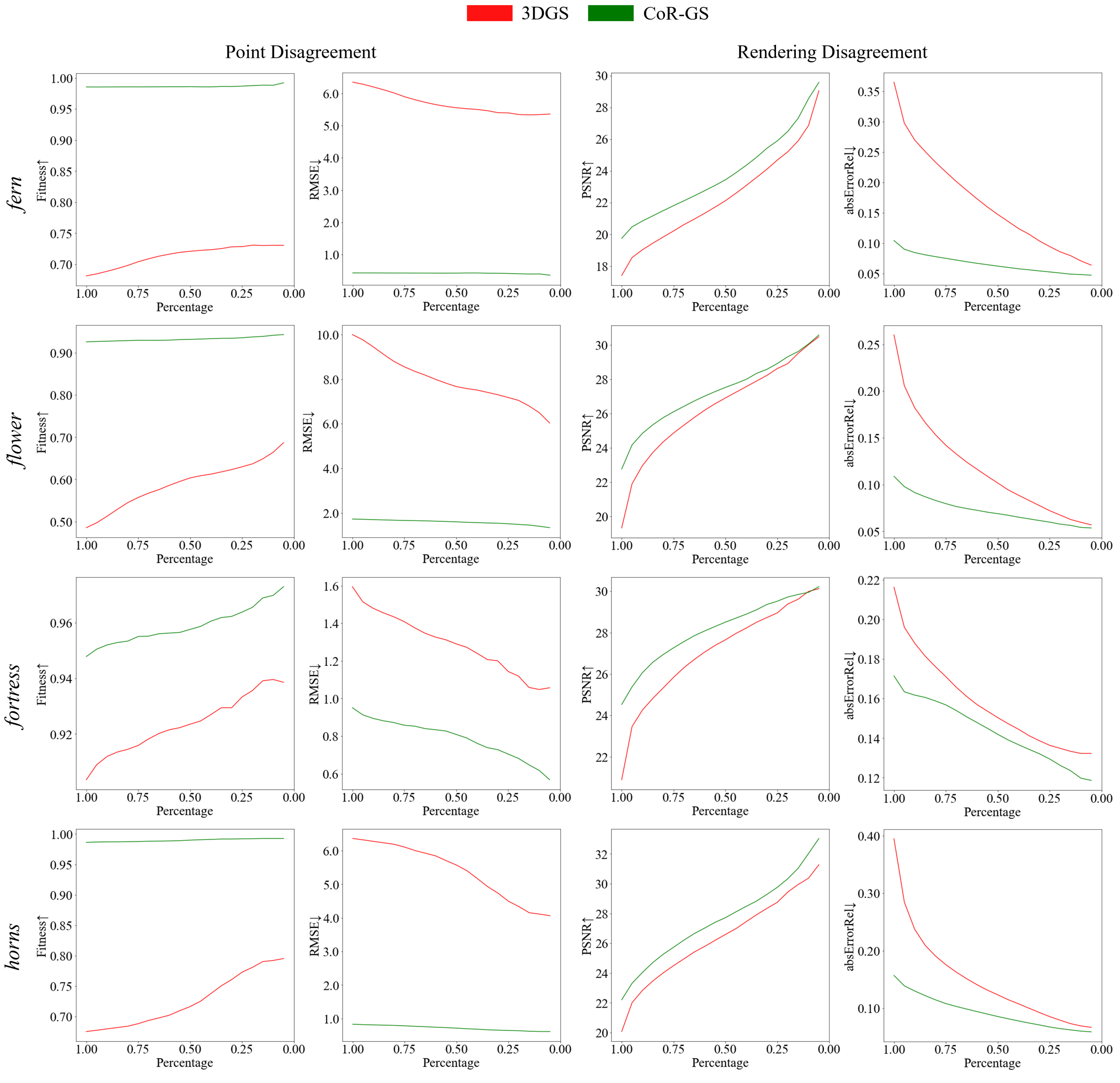}
    \caption{The correlation between the disagreements and reconstruction quality on the 3-view LLFF dataset. With the reduction of regions with higher disagreement scores, the reconstruction quality averaging the remaining regions continuously improves.}
    \label{suppl_fig:not all equal}
\end{figure}

\begin{figure}[!t]
    \centering
    \includegraphics[width=1.0\linewidth]{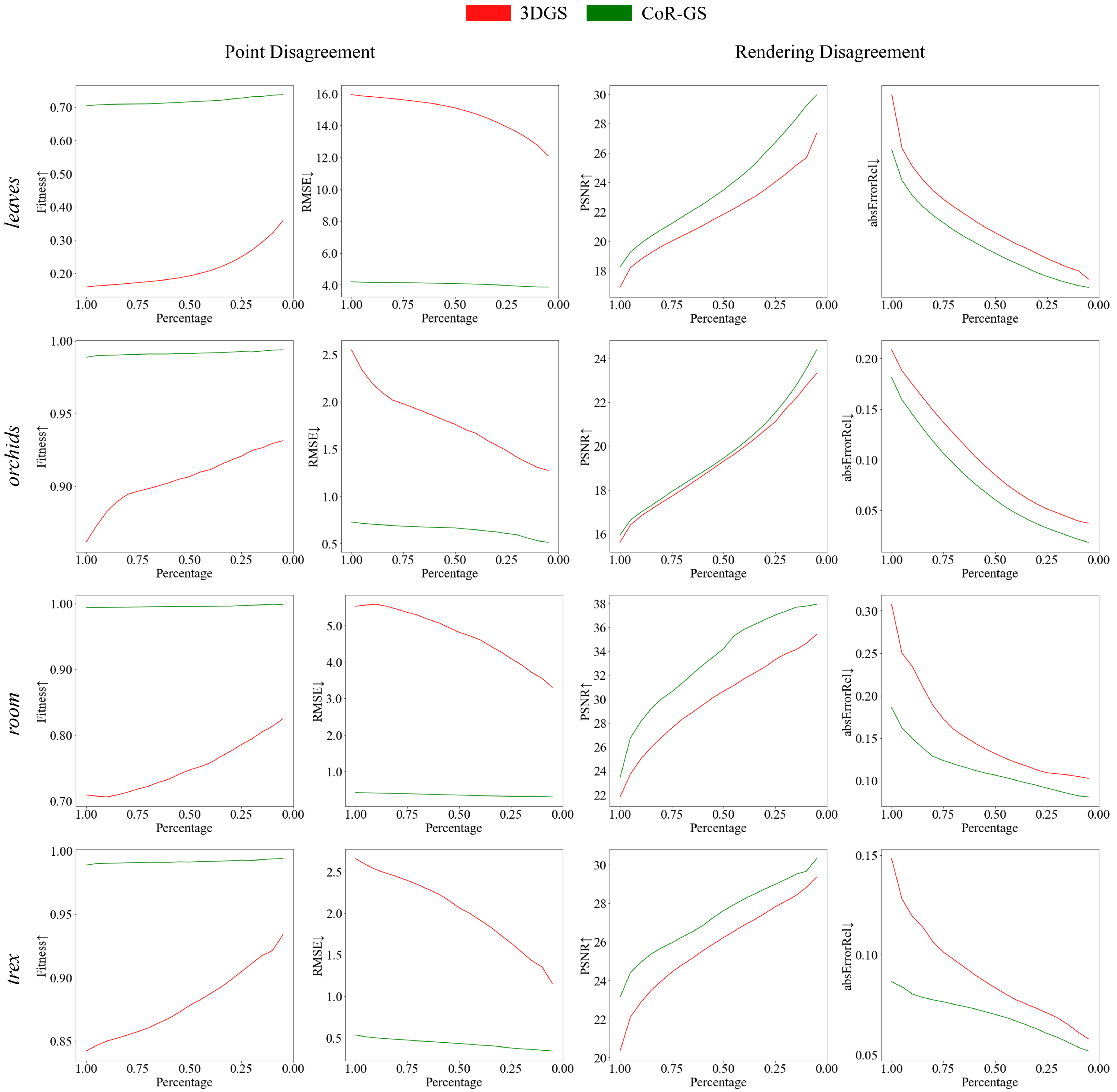}
    \caption{The correlation between the disagreements and reconstruction quality on the 3-view LLFF dataset. With the reduction of regions with higher disagreement scores, the reconstruction quality averaging the remaining regions continuously improves.}
    \label{suppl_fig:not all equal 2}
\end{figure}

\begin{figure}[!t]
    \centering
    \includegraphics[width=1.0\linewidth]{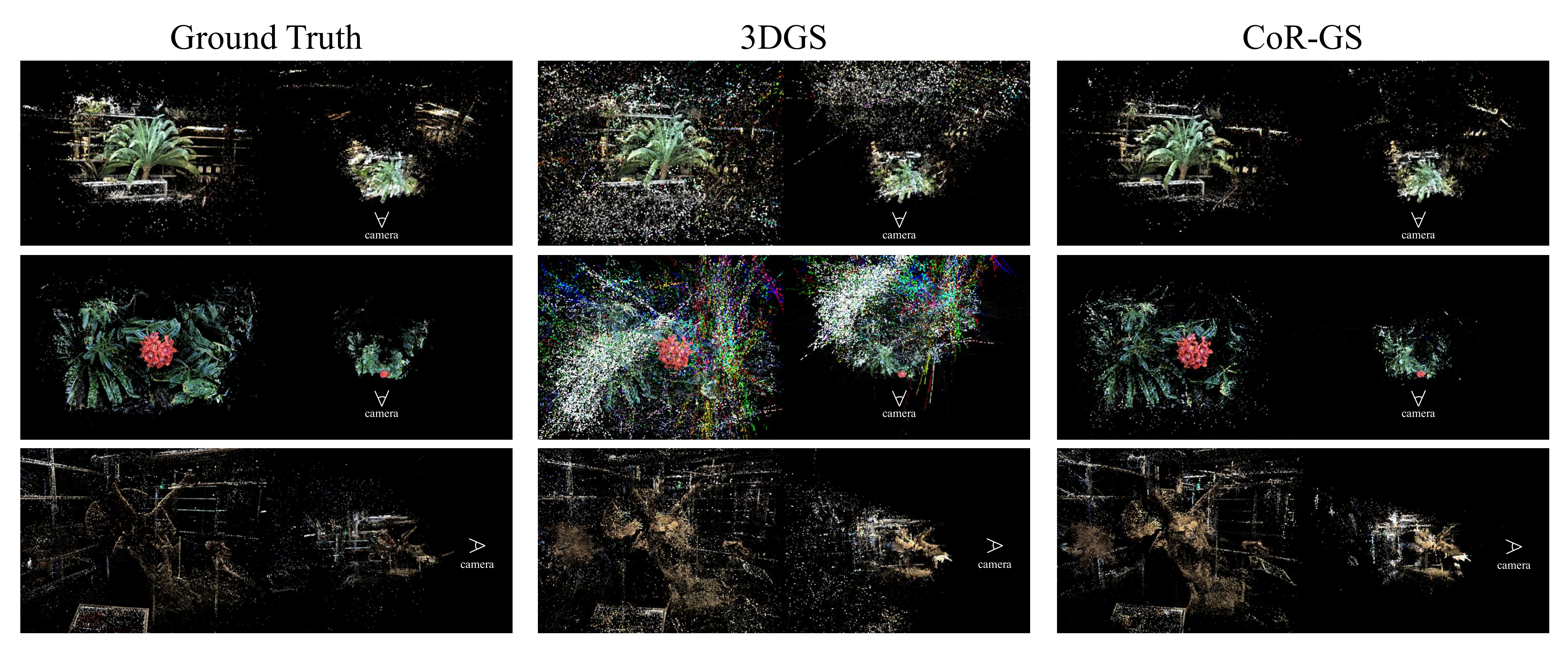}
    \caption{Visulazation of point clouds of 3D Gaussian radiance fields. We obtain the ground truth by training radiance fields with dense views. CoR-GS reconstructs more compact representations, and are more similar to dense-view representations.}
    \label{suppl_fig:point visualization}
\end{figure}

In \cref{suppl_fig:not all equal,suppl_fig:not all equal 2}, we provide results on more scenes to demonstrate the negative correlation between the disagreement and accurate reconstruction. By suppressing two disagreements, CoR-GS has an overall improvement in reconstruction quality to vanilla 3DGS, which is reflected in areas with different disagreement scores. We further provide visualization of the reconstructed Gaussian points in \cref{suppl_fig:point visualization}.

\clearpage

\section{Additional Ablation Results}
\label{suppl_sec:ablation}

We provide more ablation results of the distance threshold in co-pruning and the regularization term in pseudo-view co-regularization in \cref{suppl_tab:ablation}.

\subsection{Distance Threshold}

The distance threshold controls how co-pruning considers Gaussians as no-matching. We see the threshold $\tau=5$ and $\tau=10$ get the best results. The threshold $\tau=3$ performs worse than the final implementation $\tau=5$, which prunes Gaussians strictly. The threshold $\tau=30$ imposes very loose constraints on Gaussians, which perform similarly to solely with pseudo-view co-regularization.

\subsection{Color and Depth Co-regularization}

We also ablate the regularization term with rendered color and depth. Following the depth-regularized 3DGS \cite{zhu2023FSGS,Xiong2023sparsegs}, we compute the depth loss with the Pearson correlation. We find that although depth co-regularization alone helps, the impact becomes very weak when used in conjunction with color co-regularization. This is because color co-regularization also imposes constraints on depth information due to the sorting process implied for rendering. Therefore, we do not perform co-regularization on depth maps in this case.

\begin{table}[!t] 
  \caption{Ablation study of CoR-GS on the 3-view LLFF setting.}
  \label{suppl_tab:ablation}
  \centering
  \resizebox{0.8\linewidth}{!}{
  \setlength{\tabcolsep}{3mm}
  \begin{tabular}{@{}cc|ccc}
    \toprule
      \multicolumn{2}{c|}{Distance Threshold $\tau$}  & PSNR $\uparrow$ & SSIM $\uparrow$  & LPIPS $\downarrow$  \\
    \midrule
    \multicolumn{2}{c|}{3} &   20.36 & 0.709 & 0.198  \\
    \multicolumn{2}{c|}{5} &   20.45 & \textbf{0.712} & \textbf{0.196}  \\
    \multicolumn{2}{c|}{10} &   \textbf{20.46} & 0.711 & \textbf{0.196}  \\
    \multicolumn{2}{c|}{30} &   20.30 & 0.707 & 0.198  \\
    \bottomrule
    \toprule
    Color Co-reg. & Depth Co-reg.  & PSNR $\uparrow$ & SSIM $\uparrow$  & LPIPS $\downarrow$ \\
    \midrule
    \checkmark & & \textbf{20.45} & \textbf{0.712} & 0.196    \\
     &  \checkmark & 20.01 & 0.685 & 0.206 \\
    \checkmark &  \checkmark & \textbf{20.45} & 0.711 & \textbf{0.195} \\
  \bottomrule
  \toprule
  \multicolumn{2}{c|}{Number of Radiance Fields}  & PSNR $\uparrow$ & SSIM $\uparrow$  & LPIPS $\downarrow$  \\
    \midrule
    \multicolumn{2}{c|}{2} & 20.45 & 0.712 & 0.196 \\
    \multicolumn{2}{c|}{3} & 20.58 & 0.721 & \textbf{0.190} \\
    \multicolumn{2}{c|}{4} & \textbf{20.61} & \textbf{0.723} & \textbf{0.190}  \\
    \bottomrule
    \toprule
  \multicolumn{2}{c|}{3DGS Baseline}  & PSNR $\uparrow$ & SSIM $\uparrow$  & LPIPS $\downarrow$  \\
    \midrule
    \multicolumn{2}{c|}{FSGS \cite{zhu2023FSGS}} & 20.43 & 0.682 & 0.248 \\
    \multicolumn{2}{c|}{CoR-GS} & 20.45 & 0.712 & 0.196 \\
    \multicolumn{2}{c|}{CoR-FSGS} & \textbf{20.93} & \textbf{0.730} & \textbf{0.194}  \\
    \bottomrule
  \end{tabular}
  }
\end{table}

\subsection{Number of Radiance Fields}

The co-regularization between two 3D Gaussian radiance fields can be naturally extended to utilizing more radiance fields. In this implementation, co-pruning prunes Gaussians exhibit high point disagreement concerning any of the other radiance fields. Pseudo-view co-regularization suppresses the rendering disagreement concerning each of the other radiance fields. With more radiance fields, the disagreement reflects the inaccurate reconstruction more accurately, resulting in further improvements than using two fields.

\subsection{Incorporating with Depth Regularized 3DGS}

Equipped with our co-regularization method, 3DGS can render high-quality novel views with sparse training views. Considering depth regularization has been proven effective in the sparse view setting, we apply our co-regularization method to the depth regularized method FSGS \cite{zhu2023FSGS}. The ablation result demonstrates that our method can work well with depth regularization, and especially exhibits an advantage on the image structural SSIM score.

\section{Additional Comparison Results}
\label{suppl_sec:more comparison}

\subsection{LLFF}

We compare our method with generalized methods on LLFF in \cref{suppl_tab:llff}. The generalized methods are trained on the DTU dataset. We also provide their per-scene fine-tuned results. Our method also shows advantages to the generalized methods.

\begin{table}[!t]
  \caption{Quantitative comparisons with generalized methods on LLFF with 3, 6, 9 training views. The best are marked in red.}
  \label{suppl_tab:llff}
  \centering
  \resizebox{1\linewidth}{!}{
  \begin{tabular}{@{}l|ccc|ccc|ccc|ccc}
    \toprule
     \multirow{2}{*}{Method}   & \multicolumn{3}{c|}{PSNR$\uparrow$}  & \multicolumn{3}{c|}{SSIM$\uparrow$}   & \multicolumn{3}{c|}{LPIPS$\downarrow$}  & \multicolumn{3}{c}{AVGE$\downarrow$} \\
      &  3-view & 6-view & 9-view & 3-view & 6-view & 9-view & 3-view & 6-view & 9-view & 3-view & 6-view & 9-view \\
    \midrule
    SRF \cite{chibane2021stereo} & 12.34 & 13.10 & 13.00 & 0.250 & 0.293 & 0.297 & 0.591 & 0.594 & 0.605 & 0.313 & 0.293 & 0.296 \\
     PixelNeRF \cite{yu2021pixelnerf}  & 7.93 & 8.74 & 8.61 & 0.272 & 0.280 & 0.274 & 0.682 & 0.676 & 0.665 & 0.461 & 0.433 & 0.432 \\
     MVSNeRF \cite{chen2021mvsnerf}  & 17.25 & 19.79 & 20.47 & 0.557 & 0.656 & 0.689 & 0.356 & 0.269 & 0.242 & 0.171 & 0.125 & 0.111 \\
    \midrule
    SRF ft \cite{chibane2021stereo} & 17.07 & 16.75 & 17.39 & 0.436 & 0.438 & 0.465 & 0.529 & 0.521 & 0.503 & 0.203 & 0.207 & 0.193 \\
     PixelNeRF ft \cite{yu2021pixelnerf} & 16.17 & 17.03 & 18.92 & 0.438 & 0.473 & 0.535 & 0.512 & 0.477 & 0.430 & 0.217 & 0.196 & 0.163 \\
     MVSNeRF ft \cite{chen2021mvsnerf} & 17.88 & 19.99 & 20.47 & 0.584 & 0.660 & 0.695 & 0.327 & 0.264 & 0.244 & 0.157 & 0.122 & 0.111 \\
    \midrule
    CoR-GS (ours) & \cellcolor[HTML]{FFCCC9}20.45 & \cellcolor[HTML]{FFCCC9}24.49 & \cellcolor[HTML]{FFCCC9}26.06 & \cellcolor[HTML]{FFCCC9}0.712 & \cellcolor[HTML]{FFCCC9}0.837 & \cellcolor[HTML]{FFCCC9}0.874 & \cellcolor[HTML]{FFCCC9}0.196 & \cellcolor[HTML]{FFCCC9}0.115 & \cellcolor[HTML]{FFCCC9}0.089 & \cellcolor[HTML]{FFCCC9}0.101 & \cellcolor[HTML]{FFCCC9}0.060 & \cellcolor[HTML]{FFCCC9}0.046 \\
  \bottomrule
  \end{tabular}
  }
\end{table}

\subsection{DTU}

We compare our method with generalized methods on DTU in \cref{suppl_tab:dtu}. The generalized methods are trained on the DTU dataset. We also provide their per-scene fine-tuned results. Our method also shows advantages to the generalized methods.

\begin{table}[tb]
  \caption{Quantitative comparisons with generalized methods on DTU with 3, 6, 9 training views. The best are marked in red.}
  \label{suppl_tab:dtu}
  \centering
  \resizebox{1\linewidth}{!}{
  \begin{tabular}{@{}l|ccc|ccc|ccc|ccc}
    \toprule
     \multirow{2}{*}{Method}     & \multicolumn{3}{c|}{PSNR$\uparrow$}  & \multicolumn{3}{c|}{SSIM$\uparrow$}   & \multicolumn{3}{c|}{LPIPS$\downarrow$}  & \multicolumn{3}{c}{AVGE$\downarrow$} \\
      &   3-view & 6-view & 9-view & 3-view & 6-view & 9-view & 3-view & 6-view & 9-view & 3-view & 6-view & 9-view \\
    \midrule
    SRF \cite{chibane2021stereo}  & 15.32 & 17.54 & 18.35 & 0.671 & 0.730 & 0.752 & 0.304 & 0.250 & 0.232 & 0.171 & 0.132 & 0.120 \\
     PixelNeRF \cite{yu2021pixelnerf}  & 16.82 & 19.11 & 20.40 & 0.695 & 0.745 & 0.768 & 0.270 & 0.232 & 0.220 & 0.147 & 0.115 & 0.100   \\
     MVSNeRF \cite{chen2021mvsnerf}  & 18.63 & 20.70 & 22.40 & 0.769 & 0.823  & 0.853 & 0.197 & 0.156 & 0.135 & 0.113 & 0.088 & 0.068 \\
    \midrule
    SRF ft \cite{chibane2021stereo}   & 15.68 & 18.81 & 20.75 & 0.698 & 0.757 & 0.785 & 0.281 & 0.225 & 0.205 & 0.162 & 0.114 & 0.093 \\
     PixelNeRF ft \cite{yu2021pixelnerf}  & 18.95 & 20.56 & 21.83 & 0.710 & 0.753 & 0.781 & 0.269 & 0.223 & 0.203 & 0.125 & 0.104 & 0.090 \\
     MVSNeRF ft \cite{chen2021mvsnerf}  & 18.54 & 20.49 & 22.22 & 0.769 & 0.822 & 0.853 & 0.197 & 0.155 & 0.135 & 0.113 & 0.089 & 0.069 \\
    \midrule
    CoR-GS (ours)  & \cellcolor[HTML]{FFCCC9}19.21 & \cellcolor[HTML]{FFCCC9}24.51 & \cellcolor[HTML]{FFCCC9}27.18 & \cellcolor[HTML]{FFCCC9}0.853 & \cellcolor[HTML]{FFCCC9}0.917 & \cellcolor[HTML]{FFCCC9}0.947 & \cellcolor[HTML]{FFCCC9}0.119 & \cellcolor[HTML]{FFCCC9}0.068 & \cellcolor[HTML]{FFCCC9}0.045 & \cellcolor[HTML]{FFCCC9}0.087 & \cellcolor[HTML]{FFCCC9}0.046 & \cellcolor[HTML]{FFCCC9}0.029 \\
  \bottomrule
  \end{tabular}
      }
\end{table}

\subsection{MipNeRF-360}

More comparisons on Mip-NeRF360 are shown in \cref{tab:mipnerf}. Our method also shows advantages to the NeRF-based methods. 

\begin{table}[h]
\caption{Comparisons with NeRFs on 24-view Mip-NeRF360.}
\label{tab:mipnerf}
\centering
\setlength{\tabcolsep}{9mm}
\renewcommand{\arraystretch}{0.9}
\resizebox{1.0\linewidth}{!}{
\begin{tabular}{@{}cccc}
\toprule
Method & PSNR $\uparrow$ & SSIM $\uparrow$  & LPIPS $\downarrow$ \\
\midrule
RegNeRF \cite{niemeyer2022regnerf} & 20.55 & 0.546 & 0.398 \\
FreeNeRF \cite{yang2023freenerf} & 21.39 & 0.587 & 0.377 \\
SparseNeRF \cite{wang2023sparsenerf} & 21.43 & 0.604 & 0.389 \\
CoR-GS (ours) & \textbf{23.39} & \textbf{0.727} & \textbf{0.271} \\
\bottomrule
\end{tabular}
}
\end{table}

\clearpage

\section{Additional Visualization Results}
\label{suppl_sec:visualization}

\subsection{Comparison with NeRFs}

\begin{figure}[!t]
    \centering
    \includegraphics[width=1.0\linewidth]{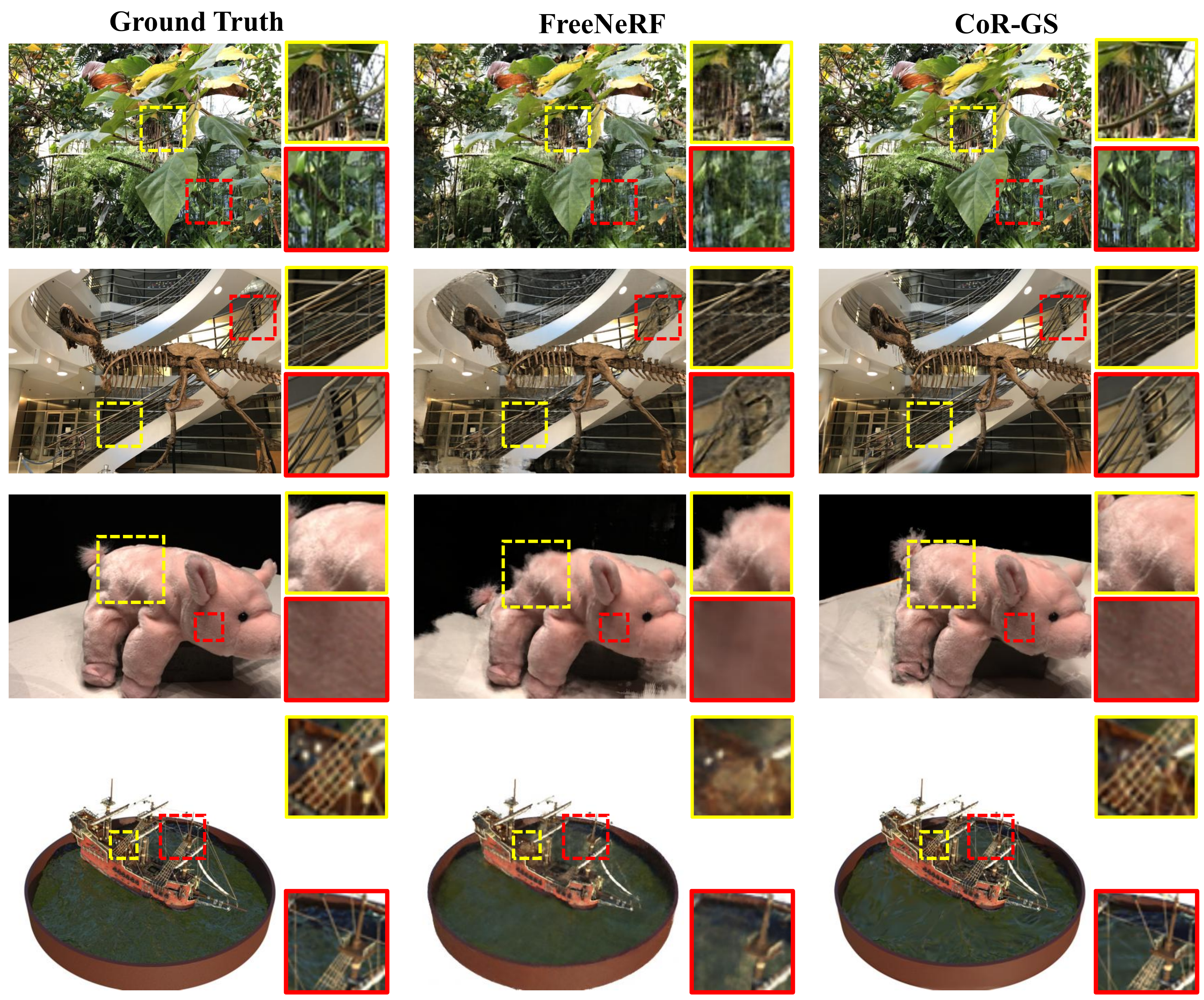}
    \caption{Qualitative comparison of FreeNeRF and CoR-GS of 3-view LLFF, 3-view DTU and 8-view Blender.}
    \label{suppl_fig:comparision with nerfs}
\end{figure}

In \cref{suppl_fig:comparision with nerfs}, we provide the visualization comparison between state-of-the-art NeRF-based method FreeNeRF \cite{yang2023freenerf} and our CoR-GS. Compared to FreeNeRF, our method demonstrates advantages in reconstructing realistic high-frequency details and better geometry of thin structures.

\subsection{CoR-GS Visualizations}

\begin{figure}[!t]
    \centering
    \includegraphics[width=1.0\linewidth]{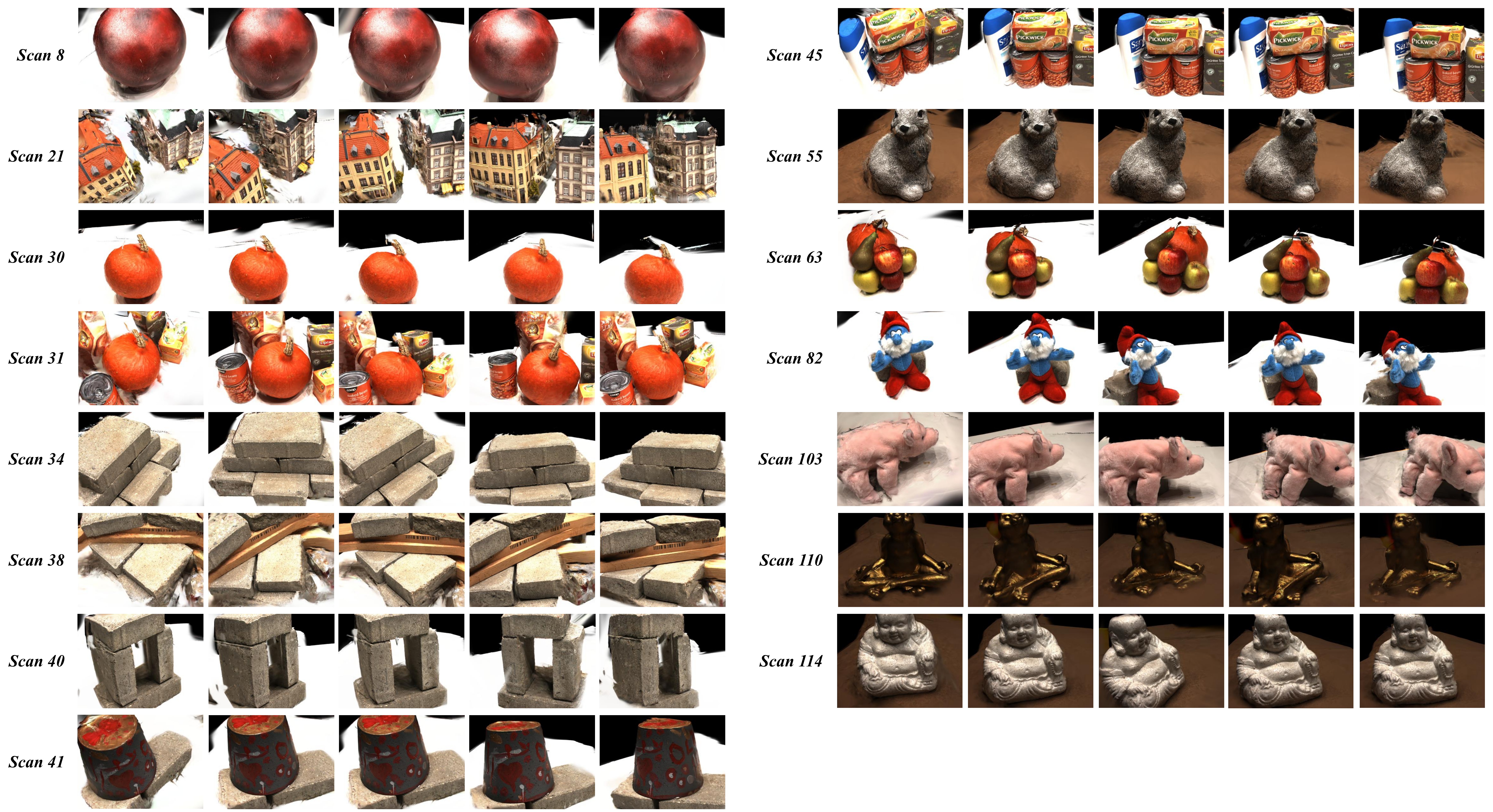}
    \caption{Qualitative Visualizations of the 3-view DTU setting.}
    \label{suppl_fig:dtu visualization}
\end{figure}

\begin{figure}[!t]
    \centering
    \includegraphics[width=1.0\linewidth]{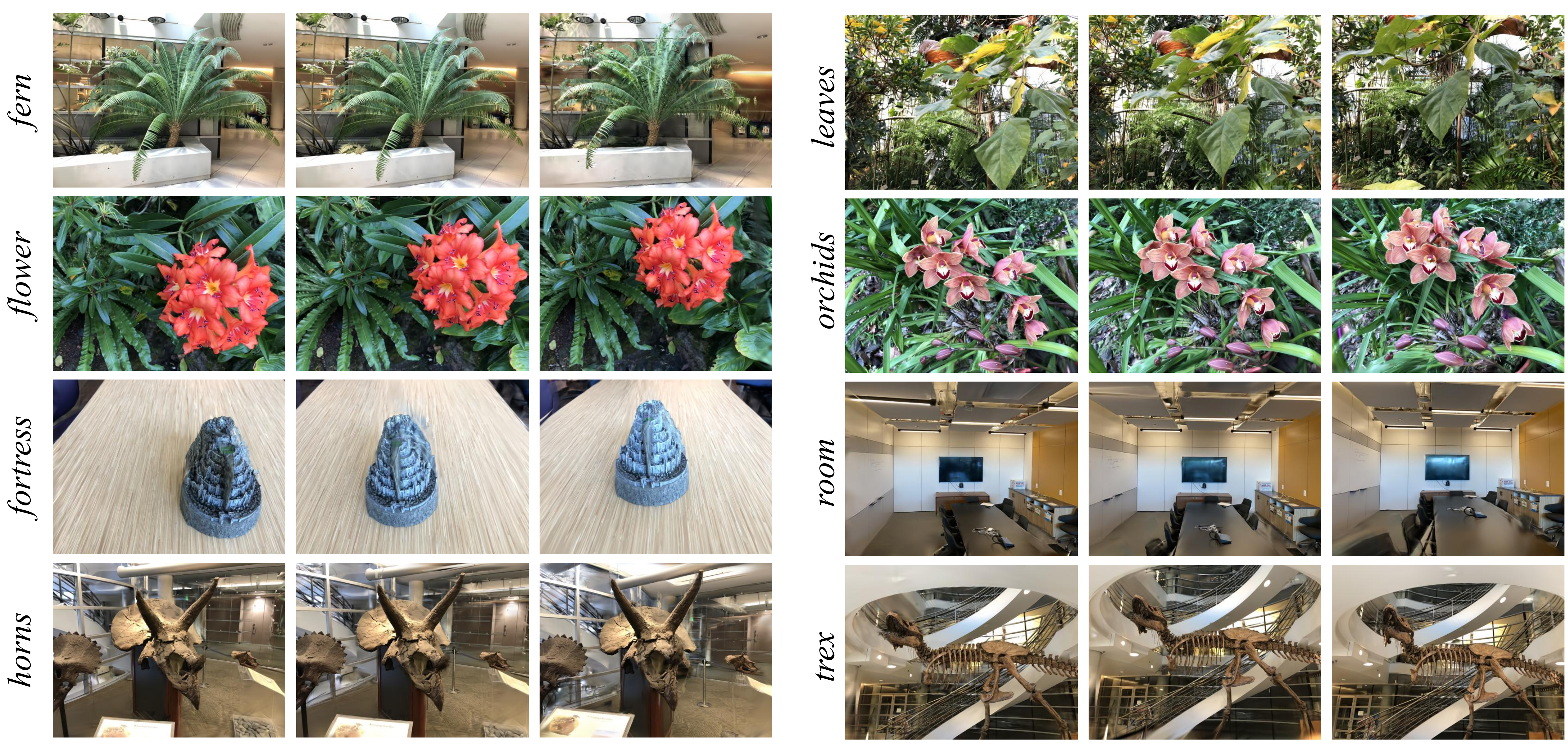}
    \caption{Qualitative Visualizations of the 3-view LLFF setting.}
    \label{suppl_fig:llff visualiaztion}
\end{figure}

\begin{figure}[!t]
    \centering
    \includegraphics[width=1.0\linewidth]{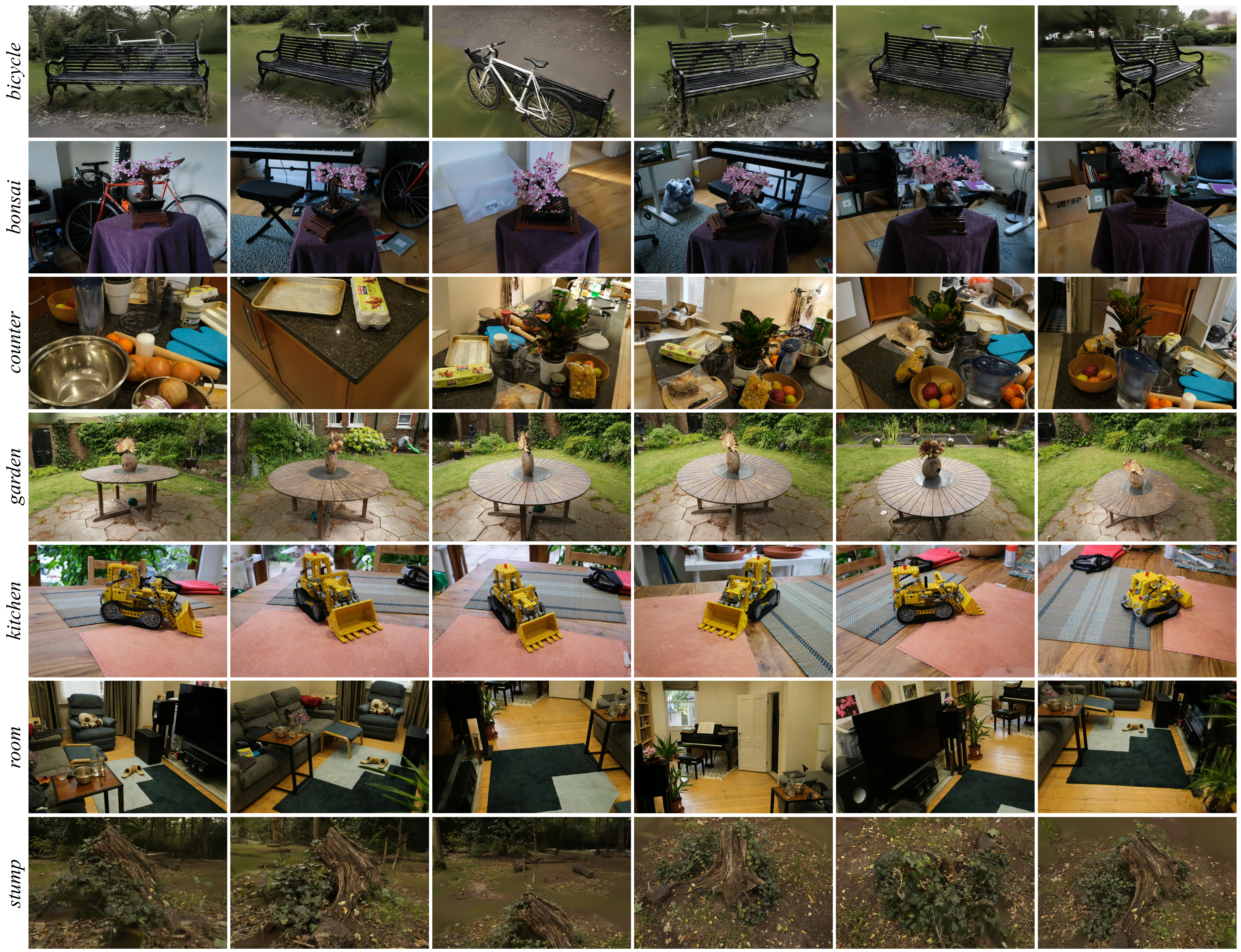}
    \caption{Qualitative Visualizations of the 24-view Mip-NeRF360 setting.}
    \label{suppl_fig:mipnerf visualiaztion}
\end{figure}

\begin{figure}[!t]
    \centering
    \includegraphics[width=1.0\linewidth]{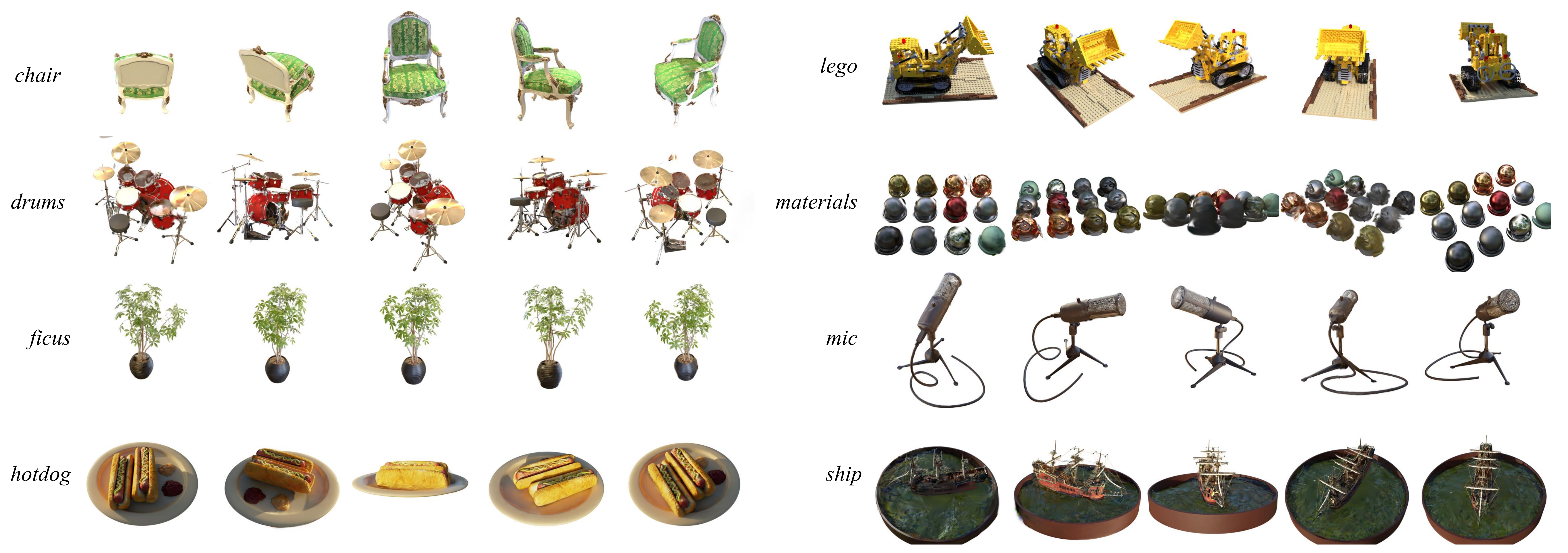}
    \caption{Qualitative Visualizations of the 8-view Blender setting.}
    \label{suppl_fig:blender visualiaztion}
\end{figure}

This section provides more visualization results of CoR-GS. \cref{suppl_fig:dtu visualization,suppl_fig:llff visualiaztion,suppl_fig:mipnerf visualiaztion,suppl_fig:blender visualiaztion} provide the novel view synthesis of CoR-GS on DTU, LLFF, Mip-NeRF360 and Blender datasets.

\section{Discussion and Future Work}
\label{suppl_sec:future}

In this paper, we propose to regularize sparse-view 3DGS from a co-regularization perspective and validate its wide effectiveness in various scenarios, with different numbers of input views and with the depth regularization method. This paper demonstrates a negative correlation between disagreed behaviors of 3D Gaussian radiance fields and the reconstruction quality in the sparse-view setting. We will investigate utilizing the disagreed behaviors in more 3DGS applications, such as video reconstruction with spatially and temporally sparse input.

\section{Details}
\label{suppl_sec:details}

\subsection{Dataset Split}
\noindent\textbf{\textit{LLFF.}}
The LLFF dataset \cite{mildenhall2019local} contains 8 forward-facing scenes. Following \cite{niemeyer2022regnerf,yang2023freenerf,wang2023sparsenerf,li2024dngaussian}, we take every 8-th image as the novel views for testing. The input views are evenly sampled across the remaining views. Images are downsampled $8\times$ to the resolution of  $378\times504$.

\noindent\textbf{\textit{DTU.}}
The DTU dataset \cite{jensen2014large} consists of 124 object-centric scenes captured by a set of fixed cameras. We follow \cite{niemeyer2022regnerf,yang2023freenerf,wang2023sparsenerf,li2024dngaussian} to evaluate models directly on the 15 scenes with the scan IDs of 8, 21, 30, 31, 34, 38, 40, 41, 45, 55, 63, 82, 103, 110, and 114. In each scan, the images with the IDs of 25, 22, and 28 are used as the input views for the 3-view setting; IDs of 25, 22, 40, 44, and 48 for the 6-view setting; IDs of 25, 22, 40, 44, 48, 0, 8, 13 for 9-view setting. The test set consists of images with IDs of 1, 2, 9, 10, 11, 12, 14, 15, 23, 24, 26, 27, 29, 30, 31, 32, 33, 34, 35, 41, 42, 43, 45, 46 and 47 for evaluation. The images are downsampled $4\times$. We use the undistorted images from COLMAP to eliminate the negative impact of unerased lens distortion.

\noindent\textbf{\textit{Blender.}}
We follow the data split used in \cite{jain2021putting,yang2023freenerf,li2024dngaussian} for the Blender dataset \cite{deng2022dsnerf}. The 8 input views are selected from the training images, with IDs 26, 86, 2, 55, 75, 93, 16, 73, and 8. The 25 test views are sampled evenly from the testing images for evaluation. All images are downsampled $2\times$ to $400 \times 400$ during the experiment.

\noindent\textbf{\textit{MipNeRF-360.}}
We follow the data split used in \cite{zhu2023FSGS} for the MipNeRF-360 dataset \cite{barron2022mip360}. We take every 8-th image as the novel views for testing. The input views are evenly sampled across the remaining views.

\clearpage

\clearpage

\bibliographystyle{splncs04}
\bibliography{main}

\begin{thebibliography}{10}
\providecommand{\url}[1]{\texttt{#1}}
\providecommand{\urlprefix}{URL }
\providecommand{\doi}[1]{https://doi.org/#1}

\bibitem{avidan1997novel}
Avidan, S., Shashua, A.: Novel view synthesis in tensor space. In: Proceedings of IEEE Computer Society Conference on Computer Vision and Pattern Recognition. pp. 1034--1040. IEEE (1997)

\bibitem{barron2021mip}
Barron, J.T., Mildenhall, B., Tancik, M., Hedman, P., Martin-Brualla, R., Srinivasan, P.P.: Mip-nerf: A multiscale representation for anti-aliasing neural radiance fields. In: Proceedings of the IEEE/CVF International Conference on Computer Vision. pp. 5855--5864 (2021)

\bibitem{barron2022mip360}
Barron, J.T., Mildenhall, B., Verbin, D., Srinivasan, P.P., Hedman, P.: Mip-nerf 360: Unbounded anti-aliased neural radiance fields. In: Proceedings of the IEEE/CVF Conference on Computer Vision and Pattern Recognition. pp. 5470--5479 (2022)

\bibitem{blum1998combining}
Blum, A., Mitchell, T.: Combining labeled and unlabeled data with co-training. In: Proceedings of the eleventh annual conference on Computational learning theory. pp. 92--100 (1998)

\bibitem{chen2022tensorf}
Chen, A., Xu, Z., Geiger, A., Yu, J., Su, H.: Tensorf: Tensorial radiance fields. In: Computer Vision--ECCV 2022: 17th European Conference, Tel Aviv, Israel, October 23--27, 2022, Proceedings, Part XXXII. pp. 333--350. Springer (2022)

\bibitem{chen2021mvsnerf}
Chen, A., Xu, Z., Zhao, F., Zhang, X., Xiang, F., Yu, J., Su, H.: Mvsnerf: Fast generalizable radiance field reconstruction from multi-view stereo. In: Proceedings of the IEEE/CVF International Conference on Computer Vision. pp. 14124--14133 (2021)

\bibitem{chen2023neurbf}
Chen, Z., Li, Z., Song, L., Chen, L., Yu, J., Yuan, J., Xu, Y.: Neurbf: A neural fields representation with adaptive radial basis functions. In: Proceedings of the IEEE/CVF International Conference on Computer Vision. pp. 4182--4194 (2023)

\bibitem{chibane2021stereo}
Chibane, J., Bansal, A., Lazova, V., Pons-Moll, G.: Stereo radiance fields (srf): Learning view synthesis for sparse views of novel scenes. In: Proceedings of the IEEE/CVF Conference on Computer Vision and Pattern Recognition. pp. 7911--7920 (2021)

\bibitem{cong2023enhancing}
Cong, W., Liang, H., Wang, P., Fan, Z., Chen, T., Varma, M., Wang, Y., Wang, Z.: Enhancing nerf akin to enhancing llms: Generalizable nerf transformer with mixture-of-view-experts. In: Proceedings of the IEEE/CVF International Conference on Computer Vision. pp. 3193--3204 (2023)

\bibitem{deng2022dsnerf}
Deng, K., Liu, A., Zhu, J.Y., Ramanan, D.: Depth-supervised nerf: Fewer views and faster training for free. In: Proceedings of the IEEE/CVF Conference on Computer Vision and Pattern Recognition. pp. 12882--12891 (2022)

\bibitem{yu2021plenoxels}
Fridovich-Keil, S., Yu, A., Tancik, M., Chen, Q., Recht, B., Kanazawa, A.: Plenoxels: Radiance fields without neural networks. In: Proceedings of the IEEE/CVF Conference on Computer Vision and Pattern Recognition. pp. 5501--5510 (2022)

\bibitem{han2018co}
Han, B., Yao, Q., Yu, X., Niu, G., Xu, M., Hu, W., Tsang, I., Sugiyama, M.: Co-teaching: Robust training of deep neural networks with extremely noisy labels. Advances in neural information processing systems  \textbf{31} (2018)

\bibitem{hu2023tri}
Hu, W., Wang, Y., Ma, L., Yang, B., Gao, L., Liu, X., Ma, Y.: Tri-miprf: Tri-mip representation for efficient anti-aliasing neural radiance fields. In: Proceedings of the IEEE/CVF International Conference on Computer Vision. pp. 19774--19783 (2023)

\bibitem{jain2021putting}
Jain, A., Tancik, M., Abbeel, P.: Putting nerf on a diet: Semantically consistent few-shot view synthesis. In: Proceedings of the IEEE/CVF International Conference on Computer Vision. pp. 5885--5894 (2021)

\bibitem{jensen2014large}
Jensen, R., Dahl, A., Vogiatzis, G., Tola, E., Aan{\ae}s, H.: Large scale multi-view stereopsis evaluation. In: Proceedings of the IEEE conference on computer vision and pattern recognition. pp. 406--413 (2014)

\bibitem{kerbl20233d}
Kerbl, B., Kopanas, G., Leimk{\"u}hler, T., Drettakis, G.: 3d gaussian splatting for real-time radiance field rendering. ACM Transactions on Graphics (ToG)  \textbf{42}(4),  1--14 (2023)

\bibitem{kim2022infonerf}
Kim, M., Seo, S., Han, B.: Infonerf: Ray entropy minimization for few-shot neural volume rendering. In: Proceedings of the IEEE/CVF Conference on Computer Vision and Pattern Recognition. pp. 12912--12921 (2022)

\bibitem{kulhanek2022viewformer}
Kulh{\'a}nek, J., Derner, E., Sattler, T., Babu{\v{s}}ka, R.: Viewformer: Nerf-free neural rendering from few images using transformers. In: European Conference on Computer Vision (ECCV) (2022)

\bibitem{li2024dngaussian}
Li, J., Zhang, J., Bai, X., Zheng, J., Ning, X., Zhou, J., Gu, L.: Dngaussian: Optimizing sparse-view 3d gaussian radiance fields with global-local depth normalization. In: Proceedings of the IEEE/CVF Conference on Computer Vision and Pattern Recognition. pp. 20775--20785 (2024)

\bibitem{li2020dividemix}
Li, J., Socher, R., Hoi, S.C.: Dividemix: Learning with noisy labels as semi-supervised learning. arXiv preprint arXiv:2002.07394  (2020)

\bibitem{liu2020nsvf}
Liu, L., Gu, J., Zaw~Lin, K., Chua, T.S., Theobalt, C.: Neural sparse voxel fields. Advances in Neural Information Processing Systems  \textbf{33},  15651--15663 (2020)

\bibitem{luiten2023dynamic}
Luiten, J., Kopanas, G., Leibe, B., Ramanan, D.: Dynamic 3d gaussians: Tracking by persistent dynamic view synthesis. arXiv preprint arXiv:2308.09713  (2023)

\bibitem{mildenhall2019local}
Mildenhall, B., Srinivasan, P.P., Ortiz-Cayon, R., Kalantari, N.K., Ramamoorthi, R., Ng, R., Kar, A.: Local light field fusion: Practical view synthesis with prescriptive sampling guidelines. ACM Transactions on Graphics (TOG)  \textbf{38}(4),  1--14 (2019)

\bibitem{mildenhall2021nerf}
Mildenhall, B., Srinivasan, P.P., Tancik, M., Barron, J.T., Ramamoorthi, R., Ng, R.: Nerf: Representing scenes as neural radiance fields for view synthesis. Communications of the ACM  \textbf{65}(1),  99--106 (2021)

\bibitem{muller2022instant}
M{\"u}ller, T., Evans, A., Schied, C., Keller, A.: Instant neural graphics primitives with a multiresolution hash encoding. ACM Transactions on Graphics (ToG)  \textbf{41}(4),  1--15 (2022)

\bibitem{niemeyer2022regnerf}
Niemeyer, M., Barron, J.T., Mildenhall, B., Sajjadi, M.S., Geiger, A., Radwan, N.: Regnerf: Regularizing neural radiance fields for view synthesis from sparse inputs. In: Proceedings of the IEEE/CVF Conference on Computer Vision and Pattern Recognition. pp. 5480--5490 (2022)

\bibitem{roessle2022dense}
Roessle, B., Barron, J.T., Mildenhall, B., Srinivasan, P.P., Nie{\ss}ner, M.: Dense depth priors for neural radiance fields from sparse input views. In: Proceedings of the IEEE/CVF Conference on Computer Vision and Pattern Recognition. pp. 12892--12901 (2022)

\bibitem{schoenberger2016sfm}
Sch\"{o}nberger, J.L., Frahm, J.M.: {Structure-from-Motion Revisited}. In: Conference on Computer Vision and Pattern Recognition (CVPR) (2016)

\bibitem{schoenberger2016mvs}
Sch\"{o}nberger, J.L., Zheng, E., Pollefeys, M., Frahm, J.M.: {Pixelwise View Selection for Unstructured Multi-View Stereo}. In: European Conference on Computer Vision (ECCV) (2016)

\bibitem{sindhwani2005co}
Sindhwani, V., Niyogi, P., Belkin, M.: A co-regularization approach to semi-supervised learning with multiple views. In: Proceedings of ICML workshop on learning with multiple views. vol.~2005, pp. 74--79. Citeseer (2005)

\bibitem{song2023darf}
Song, J., Park, S., An, H., Cho, S., Kwak, M.S., Cho, S., Kim, S.: Därf: Boosting radiance fields from sparse inputs with monocular depth adaptation (2023)

\bibitem{sun2022direct}
Sun, C., Sun, M., Chen, H.T.: Direct voxel grid optimization: Super-fast convergence for radiance fields reconstruction. In: Proceedings of the IEEE/CVF Conference on Computer Vision and Pattern Recognition. pp. 5459--5469 (2022)

\bibitem{tang2023dreamgaussian}
Tang, J., Ren, J., Zhou, H., Liu, Z., Zeng, G.: Dreamgaussian: Generative gaussian splatting for efficient 3d content creation. arXiv preprint arXiv:2309.16653  (2023)

\bibitem{Uhrig2017THREEDV}
Uhrig, J., Schneider, N., Schneider, L., Franke, U., Brox, T., Geiger, A.: Sparsity invariant cnns. In: International Conference on 3D Vision (3DV) (2017)

\bibitem{wang2023sparsenerf}
Wang, G., Chen, Z., Loy, C.C., Liu, Z.: Sparsenerf: Distilling depth ranking for few-shot novel view synthesis. In: Proceedings of the IEEE/CVF International Conference on Computer Vision (ICCV). pp. 9065--9076 (October 2023)

\bibitem{wang2004ssim}
Wang, Z., Bovik, A.C., Sheikh, H.R., Simoncelli, E.P.: Image quality assessment: from error visibility to structural similarity. IEEE transactions on image processing  \textbf{13}(4),  600--612 (2004)

\bibitem{wei2020combating}
Wei, H., Feng, L., Chen, X., An, B.: Combating noisy labels by agreement: A joint training method with co-regularization. In: Proceedings of the IEEE/CVF conference on computer vision and pattern recognition. pp. 13726--13735 (2020)

\bibitem{wu20234d}
Wu, G., Yi, T., Fang, J., Xie, L., Zhang, X., Wei, W., Liu, W., Tian, Q., Wang, X.: 4d gaussian splatting for real-time dynamic scene rendering. arXiv preprint arXiv:2310.08528  (2023)

\bibitem{wynn2023diffusionerf}
Wynn, J., Turmukhambetov, D.: Diffusionerf: Regularizing neural radiance fields with denoising diffusion models. In: Proceedings of the IEEE/CVF Conference on Computer Vision and Pattern Recognition. pp. 4180--4189 (2023)

\bibitem{Xiong2023sparsegs}
Xiong, H., Muttukuru, S.: Sparsegs: Real-time 360° sparse view synthesis using gaussian splatting (10 2023)

\bibitem{xu2022point}
Xu, Q., Xu, Z., Philip, J., Bi, S., Shu, Z., Sunkavalli, K., Neumann, U.: Point-nerf: Point-based neural radiance fields. In: Proceedings of the IEEE/CVF Conference on Computer Vision and Pattern Recognition. pp. 5438--5448 (2022)

\bibitem{yang2023freenerf}
Yang, J., Pavone, M., Wang, Y.: Freenerf: Improving few-shot neural rendering with free frequency regularization. In: Proceedings of the IEEE/CVF Conference on Computer Vision and Pattern Recognition. pp. 8254--8263 (2023)

\bibitem{yu2021plenoctrees}
Yu, A., Li, R., Tancik, M., Li, H., Ng, R., Kanazawa, A.: Plenoctrees for real-time rendering of neural radiance fields. In: Proceedings of the IEEE/CVF International Conference on Computer Vision. pp. 5752--5761 (2021)

\bibitem{yu2021pixelnerf}
Yu, A., Ye, V., Tancik, M., Kanazawa, A.: pixelnerf: Neural radiance fields from one or few images. In: Proceedings of the IEEE/CVF Conference on Computer Vision and Pattern Recognition. pp. 4578--4587 (2021)

\bibitem{yu2019does}
Yu, X., Han, B., Yao, J., Niu, G., Tsang, I., Sugiyama, M.: How does disagreement help generalization against label corruption? In: International Conference on Machine Learning. pp. 7164--7173. PMLR (2019)

\bibitem{zhang2024robust}
Zhang, J., Li, J., Huang, L., Yu, X., Gu, L., Zheng, J., Bai, X.: Robust synthetic-to-real transfer for stereo matching. In: Proceedings of the IEEE/CVF Conference on Computer Vision and Pattern Recognition. pp. 20247--20257 (2024)

\bibitem{zhang2018lpips}
Zhang, R., Isola, P., Efros, A.A., Shechtman, E., Wang, O.: The unreasonable effectiveness of deep features as a perceptual metric. In: Proceedings of the IEEE conference on computer vision and pattern recognition. pp. 586--595 (2018)

\bibitem{zhou2018open3d}
Zhou, Q.Y., Park, J., Koltun, V.: Open3d: A modern library for 3d data processing. arXiv preprint arXiv:1801.09847  (2018)

\bibitem{zhou2016view}
Zhou, T., Tulsiani, S., Sun, W., Malik, J., Efros, A.A.: View synthesis by appearance flow. In: Computer Vision--ECCV 2016: 14th European Conference, Amsterdam, The Netherlands, October 11--14, 2016, Proceedings, Part IV 14. pp. 286--301. Springer (2016)

\bibitem{zhou2023sparsefusion}
Zhou, Z., Tulsiani, S.: Sparsefusion: Distilling view-conditioned diffusion for 3d reconstruction. In: CVPR (2023)

\bibitem{zhu2023FSGS}
Zhu, Z., Fan, Z., Jiang, Y., Wang, Z.: Fsgs: Real-time few-shot view synthesis using gaussian splatting (2023)

\end{thebibliography}
\end{document}